\documentclass{ieeeaccess}
\usepackage{xcolor, soul}
\sethlcolor{yellow}
\usepackage[caption=false,font=footnotesize]{subfig}

\usepackage{booktabs} 
\usepackage{multirow}

\usepackage{cite}
\usepackage{amsmath,amssymb,amsfonts}
\usepackage{algorithmic}
\usepackage{graphicx}
\usepackage{textcomp}
\def\BibTeX{{\rm B\kern-.05em{\sc i\kern-.025em b}\kern-.08em
    T\kern-.1667em\lower.7ex\hbox{E}\kern-.125emX}}
\begin{document}
\history{Date of publication xxxx 00, 0000, date of current version xxxx 00, 0000.}
\doi{xxxxxxxxxxxxxx}

\title{Unstructured Handwashing Recognition using Smartwatch to Reduce Contact Transmission of Pathogens}

\author{\uppercase{Emanuele Lattanzi}\authorrefmark{1}, 
\uppercase{Lorenzo Calisti\authorrefmark{1}}, and 
\uppercase{Valerio Freschi\authorrefmark{1}}}

\address[1]{Department of Pure and Applied Sciences, University of Urbino, Piazza della Repubblica 13, Urbino - 61029, Italy;(e-mail: emanuele.lattanzi@uniurb.it; l.calisti@campus.uniurb.it; valerio.freschi@uniurb.it)}

\tfootnote{``This work was supported by the Department of Pure and Applied Sciences of the University of Urbino''}

\markboth
{Lattanzi E. \headeretal: Unstructured Handwashing Recognition using Smartwatch to Reduce Contact Transmission of Pathogens}
{Lattanzi E. \headeretal: Unstructured Handwashing Recognition using Smartwatch to Reduce Contact Transmission of Pathogens}

\corresp{Corresponding author: Emanuele Lattanzi (e-mail: emanuele.lattanzi@uniurb.it).}

\begin{abstract}
Current guidelines from the World Health Organization indicate that the SARS-CoV-2 coronavirus, which results in the novel coronavirus disease (COVID-19), is transmitted through respiratory droplets or by contact. 
Contact transmission occurs when contaminated hands touch the mucous membrane of the mouth, nose, or eyes so hands hygiene is extremely important to prevent the spread of the SARS-CoV-2 as well as of other pathogens. 
The vast proliferation of wearable devices, such as smartwatches, containing acceleration, rotation, magnetic field sensors, etc., together with the modern technologies of artificial intelligence, such as machine learning and more recently deep-learning, allow the development of accurate applications for recognition and classification of human activities such as: walking, climbing stairs, running, clapping, sitting, sleeping, etc. 
In this work, we evaluate the feasibility of a machine learning based system which, starting from inertial signals collected from wearable devices such as current smartwatches, recognizes when a subject is washing or rubbing its hands. 
Preliminary results, obtained over two different datasets, show a classification accuracy of about 95\% and of about 94\% for respectively deep and standard learning techniques.
\end{abstract}


\titlepgskip=-15pt

\maketitle

\section{Introduction}
\label{sec:introduction}
\begin{figure*}[!h]
\centering
\subfloat[hand washing]{\includegraphics[width=\columnwidth]{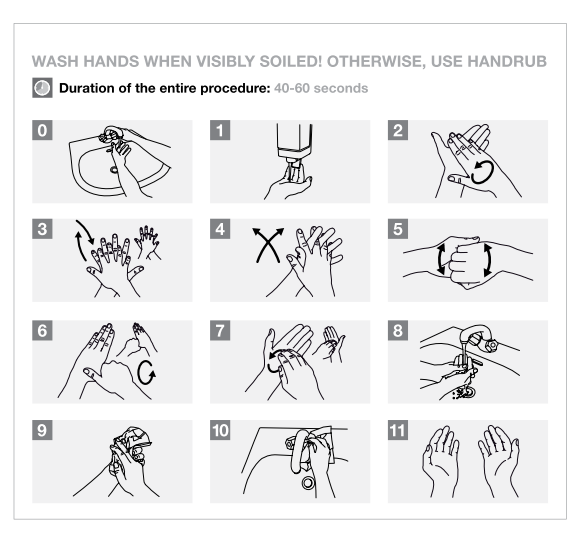}}
\subfloat[hand rubbing]{\includegraphics[width=\columnwidth]{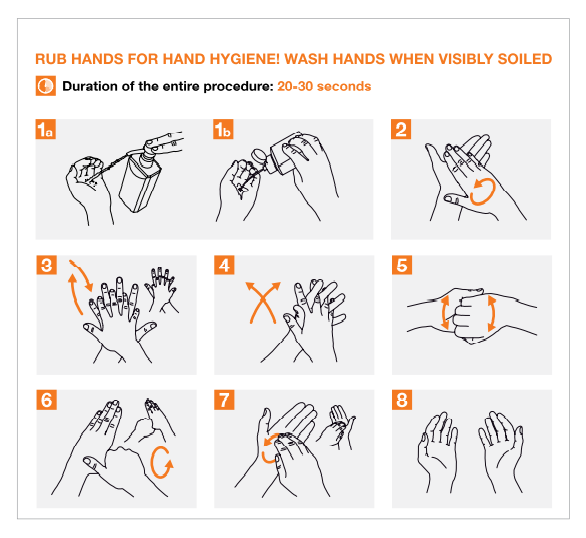}}

\caption{WHO suggested procedure to accomplish handwashing (a) and handrubbing (b).}
\label{fig:WHO_instructions}
\end{figure*}

\PARstart{T}{he} World Health Organization (WHO) indicates that hands hygiene is extremely important to prevent the transmission of bacteria and viruses by avoiding its transfer from contaminated surfaces to the mucous membrane of the mouth, nose, or eyes. Also with regard to COVID-19 disease, it is estimated that a non-negligible part of infections occurs due to contact, through our hands, with contaminated surfaces~\cite{santarpia2020}. For these reasons, one of the most important measures which any person can put in place to prevent the transmission of harmful germs is to take care of hands hygiene. To ensure proper hands hygiene, WHO suggests that one should follow either handrub, using an alcohol-based formulation, or handwash with soap and water. The two suggested procedures, described in Figure~\ref{fig:WHO_instructions}, entail different steps with different duration. In particular, water and soap handwashing comprises eleven steps and should last between 40 to 60 seconds while handrubbing only includes 8 steps with a duration of about 20 and 30 seconds. WHO also suggests that alcohol-based handrub should be used for routine decontamination of hands, while handwash with soap and water is recommended when hands are visibly soiled. 

Despite the proven effectiveness of these two procedures, most ordinary people ignore or simply do not follow them due to their non-trivial implementation. This results in a significant number of persons that limits themselves to washing/rubbing their hands as they have always been accustomed to. 

Wearable devices, such as modern smartwatches, are equipped with several sensors capable of continuously measuring characteristic parameters of our body movement. For instance, Wang et al. in 2020 have measured the accuracy of some wearable bracelets, equipped with accelerometers, gyroscopes, and electrodes for surface electromyography (sEMG), in identifying and monitoring the handwashing/handrubbing procedures suggested by WHO reaching an accuracy of over 96\% in recognizing the hands activity~\cite{Wang2020}.
 Before them, several authors have demonstrated the effectiveness of wearable devices in the classification and identification of general human activities such as running, walking, climbing/descending stairs, jumping, and sitting~\cite{zhang2013, sztyler2016, sztyler2017, bhat2018, koping2018, lattanzi2022}.

Moreover, the availability of very large data sets in human activity recognition and in medicine in general, together with the recent developments in deep learning, have led to impressive results in achieving human (or even superhuman) performance. Furthermore, current machine learning techniques have given a further boost to these studies by significantly increasing their classification accuracy which, for general macro-activities, now reaches values up to 99\%~\cite{cheng2010,singh2017,hassan2018,hou2020}. These extraordinary performances together with the availability of a large amount of sensitive data have led to the birth of new ethical issues that cannot be neglected\cite{muller2021}. 

When no WHO procedures are followed, the activity of washing one's hands is executed in a very personal way involving some totally arbitrary steps so that it can be defined as completely "unstructured". Moreover, unlike activities that involve the whole body, such as walking, running, climbing stairs, etc., handwashing involves micro-movements of highly specialized segments of the body (such as fingers), the activity of which is not taken for granted that it can be recognized starting from signal collected, for example, on the subject's wrist by means of a common smartwatch.
This high level of signals indirection together with the strong inter-subject variability which characterizes the unstructured handwashing can prevent the traditional machine learning tools from accurately recognizing it.  
For this reason, in this work, we focus on the recognition of unstructured handwashing/handrubbing with the aim to investigate the classification accuracy of an automatic smartwatch-based system capable to monitor hands hygiene in the greater part of common people. 

In particular, in this article, we introduce an extensive experimental study aimed at evaluating the ability of an automatic machine learning based system to distinguish the handwashing and handrubbing gestures from the rest of the activities that each person performs every day without the use of invasive instruments but relying only on commonly used wearable devices such as commercial smartwatches. 
Experimental results conducted on two different human activity recognition datasets across different machine learning models provide evidence of the effectiveness of this tool, which could potentially enable automatic and continuous indirect monitoring of hands hygiene of users, in an attempt to reduce the diffusion of COVID-19 and other diseases due to pathogens transmissible through direct contact.

The paper is organized according to the following structure: in Section \ref{sec:relwork} we describe state-of-the-art approaches related to our work, according to the scientific literature; in Section \ref{sec:background} we report a background description of the proposed machine learning tools; 
in Section \ref{sec:method} we illustrate the proposed method and the related design choices; in Section \ref{sec:results} we present the results of the experimental evaluation; in Section \ref{sec:conclusions} we report some conclusive remarks.

\section{Related work}
\label{sec:relwork}
At present there are no scientifically validated devices or applications that are able to recognize the activity of washing/rubbing hands by means of wearable tools.
A commercially available system called {\it SureWash}, produced by GLANTA Ltd, is able to detect the hands' movements of the hospital staff, through the use of video cameras, in order to provide information about the correct execution of the procedure defined by WHO~\cite{surewash2021}. Detecting the handwashing activity by means of video data has been investigated by several authors, for instance, Zhong et al. in 2020 presented a multi-camera system that uses video analytics to recognize seven specific actions within the hand-hygiene period~\cite{jimaging2020}. More recently, Yue et al. use a pre-trained machine learning tool for gesture recognition called YOLO v3 to successfully identify and detect the seven steps hand-washing method prescribed by several hospitals~\cite{yue2021}.
However, one of the main problems facing camera-based systems is privacy, as such systems inevitably require the installation of cameras in several rooms, and, second, it is non-ubiquitous (it is not possible to monitor self-washing/rubbing by means, for instance, of sanitizing gel). 

An orthogonal approach focuses on recognizing the handwashing activity by means of wearable inertial sensors. Here the relevant scientific contributions are reduced to a few units and most of them are based on multiple sensors with very high sensitivity and accuracy typical of scientific instrumentation~\cite{Galluzzi2015, bal2017, Li2018}. In this case, even if no video data are acquired, some privacy issues should be taken into account when designing wearable applications. In fact, several authors have shown that accelerometer data can reveal the personal identity by the way each individual moves (i.e., reveal their gender, age, and other identity markers)~\cite{jain2018, van2019}. 
From the recognition capability point of view, these preliminary works show that the automatic classification of handwashing activity, through the use of inertial sensors (accelerometers and gyroscopes), is a feasible task but, on the other hand, they do not study the potential of commercial smartwatches in common use, nor the application of modern deep-learning techniques.

Only a few relevant works which make use of commercial smartwatches have been published from 2015 to today. The first one, presented by Moldol et al. in 2015, describes a handwash monitoring and reminder system which interacts with a Bluetooth-enabled soap dispenser to recognize the start of the washing procedure~\cite{Mondol2015}.
~Thanks to these broadcast advertisements, the smartwatch can easily start processing accelerometer and gyroscope data in order to recognize each procedure step as described by WHO. 

In 2020, Mondol et al., and Banerjee et al. present two robust solutions to check the WHO compliance of the handwashing activity starting from IMU sensors signals~\cite{mondol2020, banerjee2020}.In particular they can easily identify if a component of the WHO recommendations is missing or if it is not executed as expected. More recently, also Samyoun et al. present a handwashing quality assessment system based on smartwatch~\cite{Samyoun2021}. Here the authors measure the handwashing quality in terms of likeness with the standard guidelines of WHO where the start and the end of the washing events are marked with the help of the voice interaction with the user facilitating the recognition of the activity~\cite{wu2021}.
In 2022 Wahl et al., present a detection system based on commercially available wearable device which can distinguish enacted compulsive handwashing from WHO routine~\cite{wahl2022}.

Also different combined approaches have been proposed in the last years, for instance, Wu et al. in 2021 present a prototype for autonomous hand hygiene tracking combining different IoT technologies such as IMU, video cameras, and smart dispensers which can provide prompt feedback if the handwashing is not performed as suggested by WHO. 
Finally, in 2022, Fagert et al. introduce a new sensing modality for handwashing monitoring by measuring the handwashing activity-induced vibration responses of sink structures~\cite{fagert2022}.



Although in recent years several studies concerning handwashing recognition have been proposed, none of these cover all the issues addressed in the present work. 
In particular, the contributions of the article can be summarized as follows:
\begin{itemize}

\item{we evaluate the feasibility of IMU sensors embedded on common smartwatches in recognizing the handwashing activity}
\item{we focus on unstructured handwashing rather than trying to recognize WHO suggested procedure steps}
\item{we try to distinguish between handwashing and handrubbing}
\item{we compare traditional machine learning techniques with the most modern deep learning.}

\end{itemize}

\section{Background}
\label{sec:background}
In this section, we report some background information about the machine learning tools investigated with the proposed method. In particular two {\it standard} machine learning tools and two {\it deep} learning tools have been tested. For what concerns standard learning, we evaluated Support Vector Machines (SVM) and Ensemble subspace with k-nearest neighbors (ES-KNN), while, in the deep-learning domain we have considered a Convolutional Neural Network (CNN) and a Long short-term Memory network (LSTM).   

\subsection{Ensemble subspace with k-nearest neighbors (ES-KNN)}

The k-nearest neighbors (KNN) is one of the most simple and easy to implement supervised machine learning algorithms that can be used in regression and classification problems. 
It assigns an unknown observation to the class most common among its {\it k} nearest neighbors observations, as measured by a distance metric, in the training data~\cite{fix1989, altman1992}

Despite its simplicity, KNN gives competitive results and in some cases even outperforms other complex learning algorithms. However, one of the common problems which affect KNN is due to the possible presence of non-informative features in the data which can increase miss-classification errors. This is more likely in the case of high-dimensional data. 

To improve KNN classification performances, ensemble techniques have been proposed in the literature. In general, the ensemble method entails the process of creating multiple models and combining them (for instance by averaging) to produce the desired output, as opposed to creating just one model. 
Several studies show that, frequently, an ensemble of models performs better than any individual model, because the various errors of the models average out~\cite{naftaly1997}.

One way to generate an ensemble in machine learning is to train the classifiers on different sets of data, obtained by creating several subsets from the original training set. This technique, which is commonly called {\it Ensemble subspace} has been widely explored by several authors among which the contributions of~\cite{breiman1996} and~\cite{freund1996}, which are known respectively as {\it bagging} and {\it boosting} subspace ensemble, certainly stand out.

In this work, we focus on a particular class of ensemble subspace tools applied to KNN algorithms which are called Ensemble Random Subspace KNN (ERS-KNN). According to this technique,  
the features are randomly sampled, with replacement, for each learner forcing it to not over-focus on features that appear highly predictive/descriptive in the training set, but which can fail in unknown data~\cite{ho1998,li2011}.


\subsection{Support Vector Machines (SVM)}

SVM is another class of supervised learning models traditionally used for regression and classification problems with a reduced number of samples.

An SVM model represents the input data as points in space, in such a way that the data belonging to the different classes are separated by a margin as large as possible. The new data are then mapped in the same space and the prediction of the category to which they belong is made on the basis of the side on which it falls. From the practical point of view, an SVM defines a hyperplane that best divides the dataset into the desired classes. 

Moreover, in addition to a simple linear classification, it is possible to make use of the SVM to effectively carry out nonlinear classifications using nonlinear kernel methods which implicitly map input data in a multi-dimensional feature space~\cite{steinwart2008}.


\subsection{Convolutional neural network (CNN)}

A CNN is actually a kind of multi-layer neural network following a computer vision approach to make use of any spatial or temporal information in the data. The CNN, in fact, was inspired by the biological process that occurs in the animal visual cortex, where neurons handle responses only from separate regions of the visual field. In the same way, CNN makes use of convolving filters to handle local regions within the data.
A CNN is mainly composed of an input layer, several convolutional layers, pooling layers, and fully connected layers. 
The input layer has the task of collecting data and of forwarding it to the subsequent layer. The convolutional layer represents the main core of a CNN as it contains several convolution filters, called kernels, which convolve with the input data. The operation of convolution automatically extracts useful features from the input data and reduces its dimension. Moreover, the pooling layer, also called subsampling layer, is also inserted to further reduce the number of parameters and the resulting computational cost. It includes max-pooling and/or average-pooling operations which sample, respectively, the max and the average value from the input. Finally, one or more fully connected layers act as a traditional Perceptron network which takes as input the features that originated from the previous layer. 

A CNN is traditionally built using several layer batteries and it is used in the deep-learning approach also thanks to its characteristic of eliminating the requirement of feature extraction and feature selection often at the cost of an increase in computational complexity and memory usage~\cite{Albawi2017}.

\subsection{Long short-term memory (LSTM)}
Long short-term memory (LSTM) is an artificial recurrent neural network (RNN) architecture used in the field of deep learning. LSTM networks are mostly designed to recognize patterns inside sequences of data such as numerical time series. RNN and LSTM differ from classical artificial neural networks due to the fact that they have a temporal dimension and they can not only process single data points, such as images, but also entire sequences of data such as speech or video.

A common LSTM unit is composed of a cell, an input gate, an output gate and a forget gate. The cell remembers values over arbitrary time intervals and the three gates regulate the flow of information into and out of the cell. It also makes decisions about what to store, and when to allow reads, writes, and erasures, via gates that open and close~\cite{hochreiter1997}.

LSTM networks are well-suited to classifying, processing, and making predictions based on time series data, and they have been used in many complex problems such as handwriting recognition, language modeling and translation, speech synthesis, audio analysis, and protein structure prediction, and many others~\cite{yu2019}.

\section{The proposed method}
\label{sec:method}
In this work, we evaluate the suitability of four different supervised classification methods, namely SVM, ERS-KNN, CNN, and LSTM, for classifying handwashing and handrubbing activities starting from gyroscopic and accelerometer data sampled in real-life conditions by means of common smartwatches. 

\subsection{Experimental protocol and data gathering}
\label{sec:protocol}

The classification accuracy of the proposed machine learning models has been evaluated on top of two different datasets which are an ad-hoc collected dataset and the Daily Living Activities (DLA) dataset~\cite{leotta2021,leotta2021data}.

The need for ad-hoc collecting data is due to the fact that to date there is no publicly available dataset for handwashing and handrubbing quality assessment through accelerometers and gyroscopic signals. For this reason
, we collected sensors data from a wearable Inertial Measurement Unit (IMU) positioned on the wrist of the dominant hand of four participants during real-life activities. 
In particular, each subject was asked to annotate the start and the end of each handwashing or handrubbing activity performed during the day. Each subject was wearing the IMU sensor for several hours during different days leading to a total of about 40 hours of recording containing about 1 hour and 40 minutes of total time spent washing hands and about 2 hours and 10 minutes of time spent in rubbing. The wearable device was programmed to sample its triaxial accelerometer and gyroscope at a frequency of 100 Hz and to store the collected data on the internal SD card. 
In order to remove sensors bias, the device was calibrated once at the start of the study by placing it on a stable surface and the accelerometers and gyroscopic measurements were recorded for 30~seconds.

Notice that the subjects were not instructed on how to wash or rub their hands leaving them completely free to use their usual way so to collect data about the unstructured way people normally use to wash their hands. Table~\ref{tb:db_info} shows the average duration, together with the standard deviation, of each activity performed by the four subjects. 

\begin{table}[htbp]
  \caption{Recorded activities duration in seconds.}
  \centering
    \begin{tabular}{ccc}
    \toprule
    \textbf{subject } & \textbf{handwashing } & \textbf{handrubbing} \\
\midrule
    0     & $66.68s \pm 18.69s$ & $23.66s \pm 6.26s$ \\
    1     & $31.92s \pm  8.97s$ & $26.09s \pm 3.67s$ \\
    2     & $39.47s \pm  8.52s$ & $19.18s \pm 4.29s$ \\
    3     & $30.54s \pm  6.17s$ & $25.44s \pm 8.59s$ \\
    \midrule
    \textbf{avg} & $\mathbf{50.92 \pm 22.29}$ & $\mathbf{23.59 \pm 7.33}$ \\
    \bottomrule
    \end{tabular}%
  \label{tb:db_info}%
\end{table}%
 
As we collect data in an unstructured way, the average duration and repeatability of each activity significantly depend on the subject. If this can represent an advantage in recognizing a particular subject, since her/his way of washing the hands could represent a kind of fingerprint, it could also represent a problem by reducing the ability to generalize the true activity recognition.

To collect data about daily activities we use a Shimmer3 IMU unit equipped with triaxial accelerometers and gyroscope~\cite{Shimmer3}. 
This unit is a reference prototype designed for wearable applications frequently used in activity monitoring and sports science and it is representative of the IMU family that currently equips commercial smartwatches.

The internal accelerometer is a wide range sensor sampled at 14 bits which can be configured to operate in a full-scale range from $\pm 2.0$ $g$ up to $\pm 16.0$ $g$ with a resulting sensitivity from $1~mg/LSB$ in the $\pm 2.0$ $g$ range up to $12~mg/LSB$ in the $\pm 16.0$ $g$ range typical of a sensor which most smartwatches are equipped with.


  


Finally, in order to further evaluate the ability to generalize of the proposed classification models we conduct a set of experiments starting from the DLA dataset which is one of the few public datasets available that contains handwashing data sampled through inertial sensors.  
This dataset is based on samples recorded from different parts of the body and using different wearable sensors. In particular, data are recorded on the wrist, hip, and ankle while 8 healthy volunteers, aged between 23-37, perform 17 different daily-life activities but, since our purpose is to evaluate the ability to classify handwashing through a smartwatch, in our experiments, we have used only the data collected on the person's wrist.
For the same reason, the activities that involve the whole body, such as walking, upstairs, downstairs, etc, have been removed so that only the "hands activities" such as handwriting, handwashing, sweeping, etc., have been used for our experiments. 


Notice that, this dataset does not contain data sampled during handrubbing and, moreover, the sensor placed on the person's wrist was not equipped with a gyroscope. 
Thanks also to these differences, this experiment can be very useful in demonstrating the generalization capabilities of the proposed models and in evaluating the usefulness of the signals coming from the gyroscope in handwashing recognition.

\subsection{Signal windowing} 

The recorded tracks, composed of six distinct signals (i.e. 3 accelerometer and 3 gyroscope waveforms) for the ad-hoc dataset, and of only three accelerometer signals in the case of DLA dataset, have been divided into time windows and each of these has been considered as a sample to be used to train and test the classifiers. 
Furthermore, each sample has been labeled using the annotations provided by each subject in accordance with the following three categories: {\it washing, rubbing}, and {\it other}, for the first dataset, and with the following ten {\it handwriting, handwashing, facewashing, teethbrush, sweeping, vacuuming, eating, dusting, rubbing,} and {\it other} in the case of DLA dataset.

Obviously, deciding the size of the time window is a non-trivial task because it can influence the performance of classification models in different ways.
In fact, it must be large enough to capture the "fingerprint" of the particular activity that we want to recognize, but it must not be too large to include consecutive activities. 
For what concerns human activity recognition (HAR), different window lengths have been used in the literature: starting from  1s up to 30s~\cite{cheng2010, hassan2018,hou2020}.
In particular, for what concerns the handwashing recognition both~\cite{Li2018} and~\cite{Samyoun2021} use a very tiny time window (only 0.06 seconds), with 70\% overlap between subsequent windows, due to the fact that they aim at recognizing each step of the structured handwashing procedure.~
In 2015, both~\cite{Galluzzi2015} and~\cite{Mondol2015} use a larger window respectively of 0.5 and 1 second while~\cite{Wang2020} found that a window with 0.2 seconds of amplitude and 75\% overlap gave the best classification accuracy.
~To better highlight the impact of the window length on the overall classification performance, we present in this work an extensive sensitivity analysis of the classifiers with respect to this parameter.  

Notice that, due to the proposed gathering protocol, which plans to continuously record sensors data during real-life activity, the number of samples containing non washing/rubbing events is much greater than that which contains them. 
For this reason, the samples labeled as {\it other} have been randomly undersampled in order to rebalance the occurrence of each class.

\subsection{The classifiers}
\label{sec:tools}

As machine learning classifier models we used multi-class SVM and ERS-KNN, CNN and LSTM (for neural networks). 
For what concerns the SVM tool, a cubic polynomial kernel has been choosen for performance reasons~\cite{steinwart2008}. We also considered other kernels (i.e. linear, quadratic, or Gaussian functions), however, these did not reach the performances of the cubic kernel.

\begin{figure*}[!t]
\centering
\subfloat[CNN architecture]{\includegraphics[width=1.8\columnwidth]{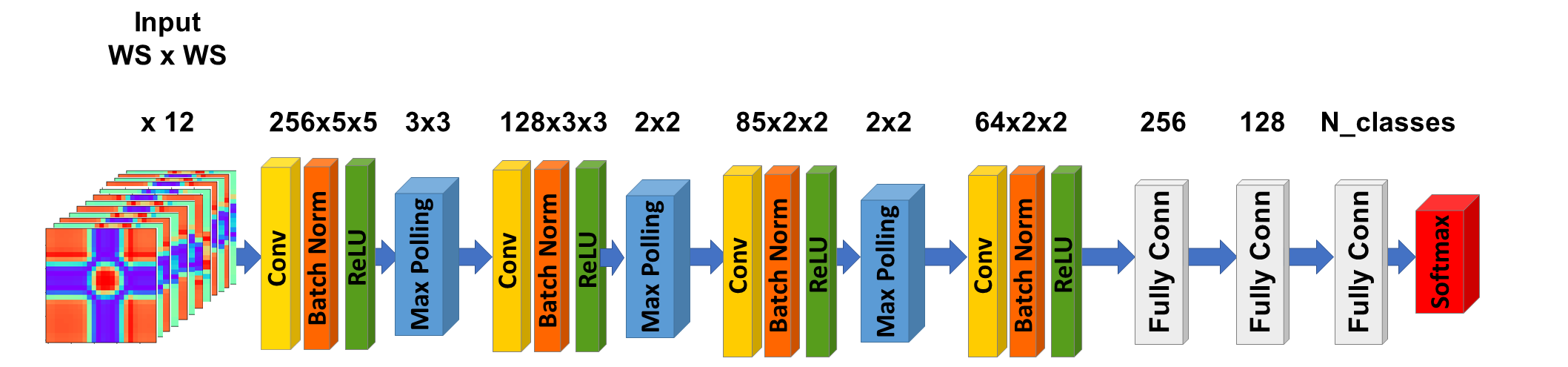}}\\
\subfloat[LSTM network architecture]{\includegraphics[width=1.8\columnwidth]{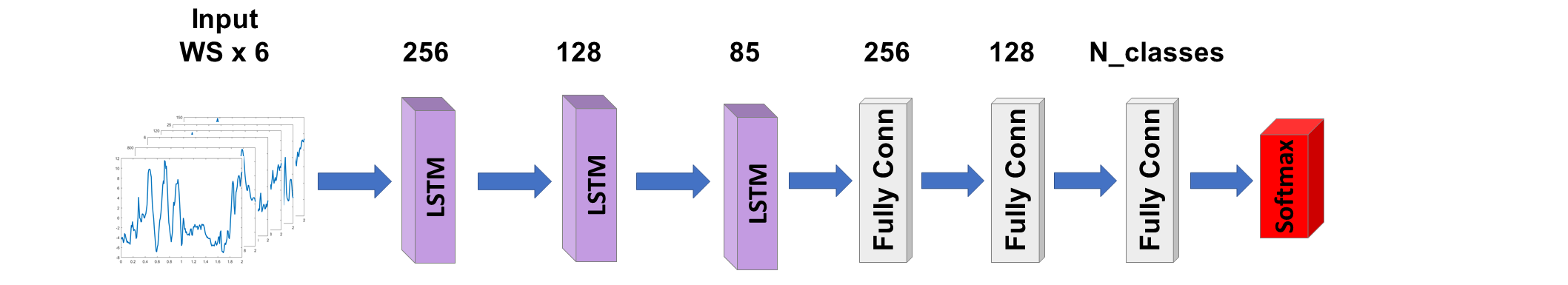}}

\caption{The architecture of proposed deep learning networks.}
\label{fig:networks_architecture}
\end{figure*}


In order to train and test the classification accuracy of standard (i.e. not based on neural network approaches) machine learning tools, the input signal needs to be processed to extract synthetic features. In particular, in this work, for each window three sets of descriptors have been computed. 
The first set, hereafter referred to as {\it Base}, contains basic statistical descriptors aimed at capturing data tendency and variability. 
These are the following classical descriptors: {\it i)} average; {\it ii)} maximum value; {\it iii)} standard deviation; {\it iv)} median value.
The second set contains the so called {\it Hjorth} parameters which are:  i) {\it activity}; ii) {\it mobility}; iii) {\it complexity}. 
Finally, the last set is built with {\it Kurtosis} and {\it Skewness} parameters aimed at capturing the {\it Shape} of the data. 

While the {\it Base} set easily describes the sample tendency, {\it Hjorth} parameters can capture the main characteristics of the signal in the frequency domain. In fact, {\it Hjorth} activity represents the power of the signal, the mobility its mean frequency, and the complexity measures its change in frequency~\cite{Hjorth1970}.

Kurtosis and Skewness are used to describe, respectively, the degree of dispersion and symmetry of the data. In particular, Kurtosis is a measure of whether the data are heavy-tailed or light-tailed relative to a normal distribution while Skewness measures how much data differ from a completely symmetrical distribution~\cite{kim2004}.

The architecture of the CNN and LSTM networks are presented, respectively in Figure~\ref{fig:networks_architecture}.(a) and Figure~\ref{fig:networks_architecture}.(b).
In the case of deep learning approaches no feature extraction is needed and the samples of the signals, that make up the time window, can be directly used as input for the classification tool.
In the case of CNN, which is commonly applied to analyzing visual imagery and which has been designed to work properly with bidimensional data, a preprocessing step has been added to represent time series data by means of visual cues. This possibility has recently attracted widespread attention so that in literature we can count several strategies aimed at re-coding time series into images to enable computer vision techniques and to perform classification~\cite{Wang2015,baldini2017,qin2020}.
In this paper, in particular, we investigate the method proposed by~\cite{Wang2015} which encodes time series as images called Gramian Angular Summation/Difference Field (GASF/GADF). This method represents time series in a polar coordinate system instead of the typical Cartesian coordinates with the advantage of preserving spatial and temporal relations. 
Because this method leads to the production of two distinct images: one for the  Gramian Angular Summation (GASF) and one for the Gramian Angular Difference (GADF), in the case of the ad-hoc dataset, we obtain 12 images (six from the accelerometer and six from the gyroscopic data), which reduce to only six for DLA dataset.
 
As a consequence, the CNN model takes in input a 12-channel or a 6-channel square image, depending on the dataset used, reconstructed starting from these data, whose height and width depend on the chosen window processing size ($WS$). 
The image is then convolved by four subsequent convolutional layers with decreasing size and numbers of filters. Furthermore, each convolutional layer is followed by a batch normalization layer and by a rectified linear activation function ($ReLu$).
Batch normalization is used to standardize the input before forwarding it to the following layer and it has the effect of stabilizing the learning process and reducing the number of training epochs required to train the network~\cite{ioffe2015}.    
The results of the ReLu layer are then processed by a subsequent pooling layer which selects the most activated features ({\it max pooling}). At the end of the convolutional structures, three fully connected layers, with decreasing number of neurons, have been added. The output of the last layer is then processed by a softmax function which assigns to each class a probability proportional to the output signal.  

The LSTM network, on the other hand, receives in input six (three for DLA) sequences extracted from the original time series the length of which is the size of the window processing ($WS$). The input is then processed by three subsequent LSTM layers with decreasing number of hidden units. The output of the last LSTM layer is then forwarded to three fully connected layers as in the case of CNN.

\subsection{Classification performance metrics}

For the proposed classifiers we calculate several performance metrics, together with the standard deviations, during a k-fold cross-validation test with k=5. 
Dealing with multi-class classifiers, entails the evaluation of the following quantities for each of the {\it N} classes ($i\in[1\cdots N]$ is an index that identifies a specific class): 
$TP_i$, the number of true positives predicted for class $i$; 
$TN_i$, the number of true negatives predicted for class $i$;
$FP_i$, the number of false positives predicted for class $i$; 
$FN_i$, the number of false negatives predicted for class $i$.

Subsequently, these indicators have been used to compute the following metrics (corresponding to the so called {\it macro-averaging} measures) \cite{sokolova2009}: 
\begin{equation}
\label{eq:precision}
	Precision = \frac{1}{N}\sum_{i=1}^{N}\frac{TP_i}{TP_i + FP_i}
\end{equation}  

\begin{equation}
\label{eq:recall}
Recall = \frac{1}{N}\sum_{i=1}^{N}\frac{TP_i}{TP_i + FN_i}
\end{equation}

\begin{equation}
\label{eq:f1score}
F1 score = 2 \cdot \frac{Precision \cdot Recall}{Precision + Recall}
\end{equation}

\begin{equation}
\label{eq:accuracy}
Accuracy = \frac{1}{N}\sum_{i=1}^{N}\frac{TP_i + TN_i}{TP_i + TN_i + FP_i + FN_i}
\end{equation}

\section{Results and Discussion}
\label{sec:results}

In this section, we report and discuss the obtained results. 
First of all, we show the best classification metrics calculated with the four machine learning tools described in section~\ref{sec:tools} on the ad-hoc built dataset. Then, for each classifier, we report the sensitivity analysis with respect to the processing window length and the feature selection results for SVM and ERS-KNN.

Finally, in the last part of the section, we show the results obtained while running two sets of experiments starting from the publicly available DLA dataset.  

\subsection{Classification results}
\label{sec:classification_res}

In this section, the results related to the classification of the human activity and of the subject identity obtained on the ad-hoc built dataset are reported. 

\subsubsection{Human activity classification}
\label{sec:activity_classification_ad_hoc}

\begin{table}[htbp]
  \caption{Best activity classification results obtained with the proposed models.}
  \centering
    \begin{tabular}{rcccc}
    \toprule
          & SVM & ERS-KNN & LSTM & CNN \\
    \midrule
    Accuracy    & $0.942$ & $0.946$ & $\mathbf{0.947}$ & $0.909$ \\
    Precision   & $0.936$ & $\mathbf{0.941}$ & $0.911$ & $0.898$ \\
    Recall      & $\mathbf{0.934}$ & $0.932$ & $0.908$ & $0.917$ \\
    F1-score    & $0.935$ & $\mathbf{0.936}$ & $0.910$ & $0.908$ \\
    \bottomrule
    \end{tabular}%
  \label{tab:activities_classification}%
\end{table}%

Table~\ref{tab:activities_classification} reports the best value of the classification metrics obtained when using the four proposed models. These values refer to the higher results obtained for each model when varying the window processing size and, for the standard learning tools, also the number of selected features. Each value is reported as the average value 
calculated during the 5-fold cross-validation test. For each metric, the highest value obtained ever is highlighted in bold. For instance, the SVM classifier obtains the highest Recall value (about $0.934$) while the ERS-KNN shows the highest Precision and F1-score (respectively $0.941$ and $0.936$). LSTM, on the other hand, reaches the best accuracy value of about $0.947$. This suggests that the classification of the handwashing/handrubbing activities using signals gathered from a common smartwatch is a well feasible task which can be accomplished both with standard or deep learning techniques. 
Notice that, the best results reported here have been obtained with the following size of the processing window: SVM=12s; ERS-KNN=8s; LSTM=2s; CNN=6s. Moreover, in the case of SVM and ERS-KNN tools, all the proposed features have been used.

\begin{figure*}[!h]
\centering
\subfloat[SVM]{\includegraphics[width=0.9\columnwidth]{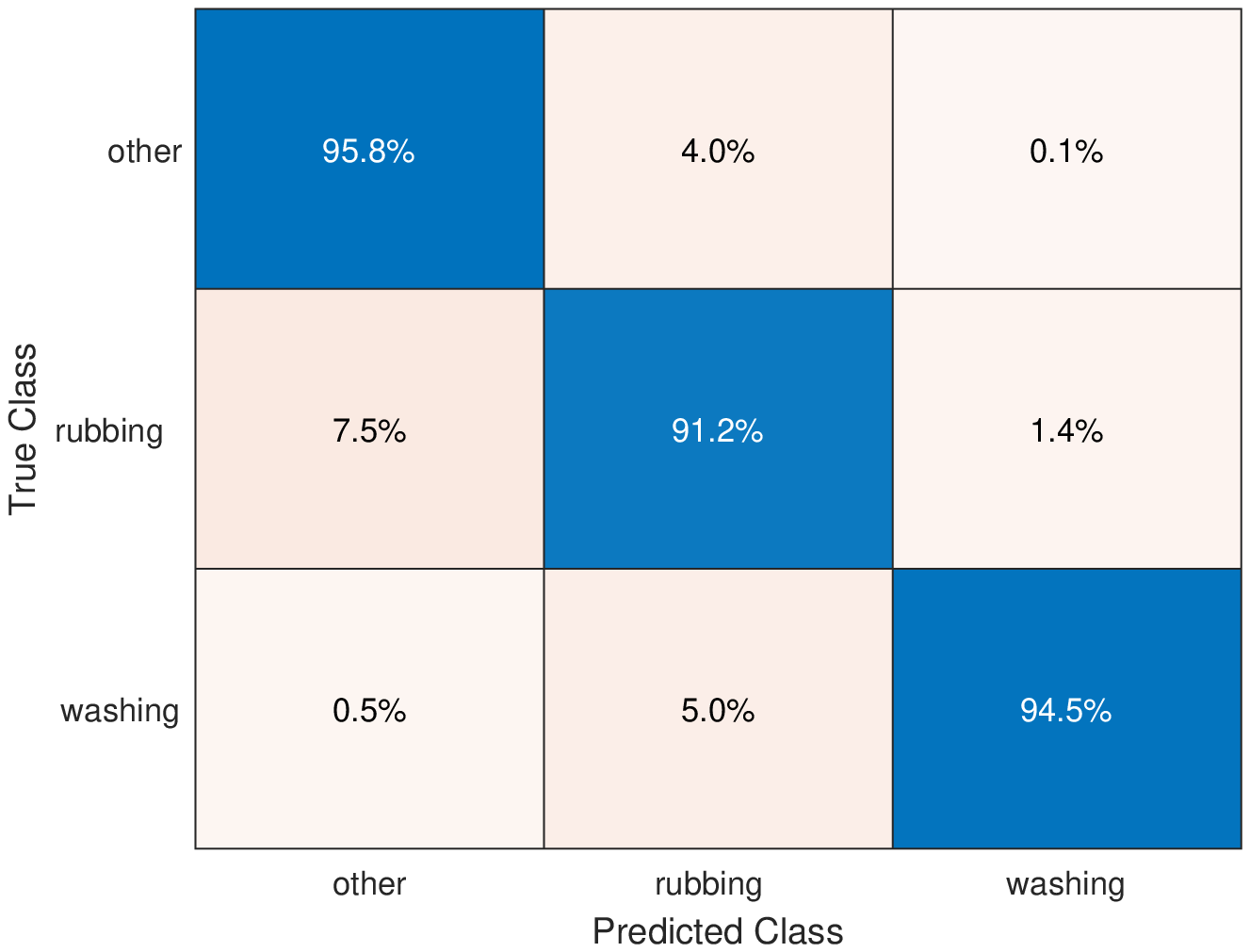}}
\subfloat[ERS-KNN]{\includegraphics[width=0.9\columnwidth]{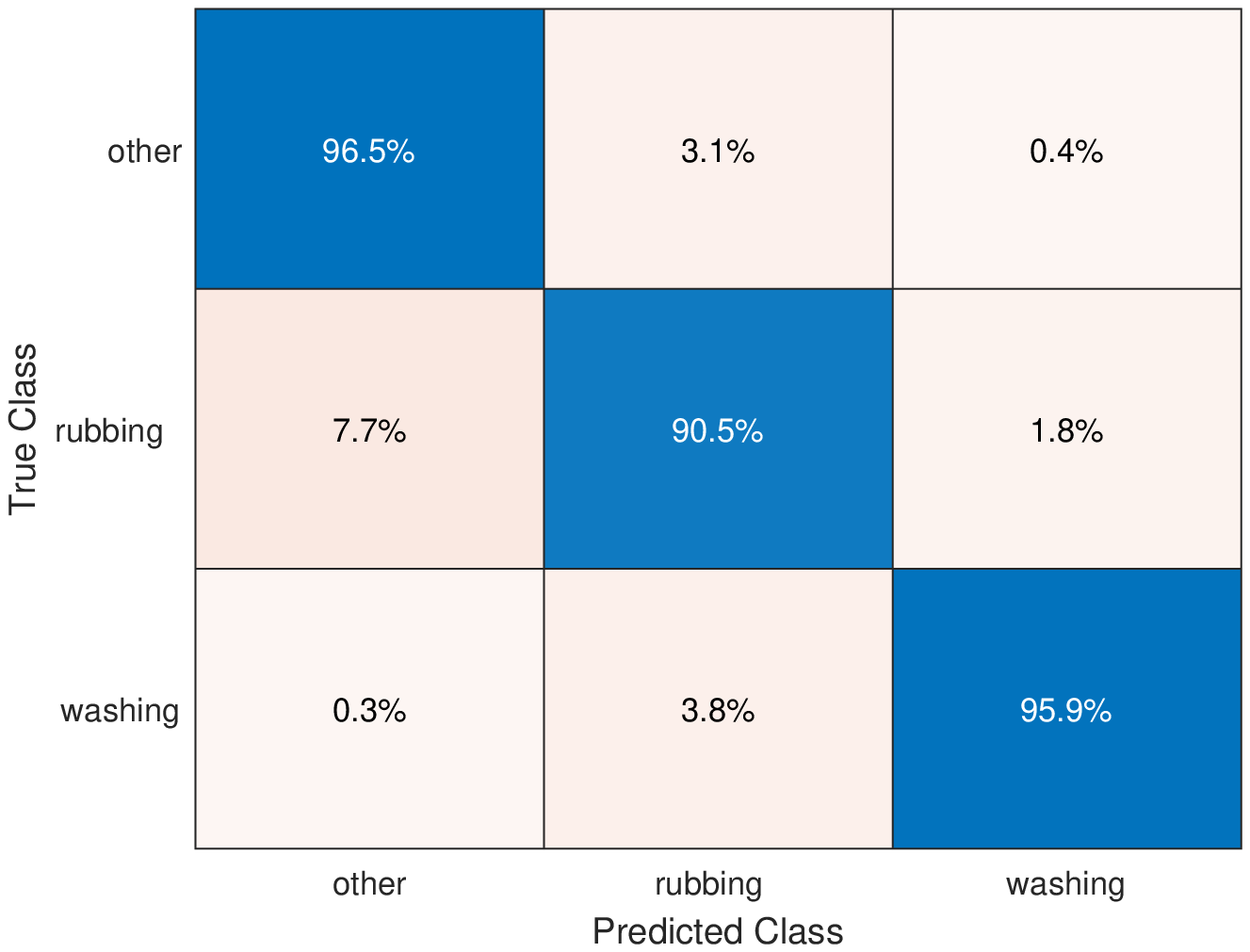}}\\
\subfloat[LSTM]{\includegraphics[width=0.9\columnwidth]{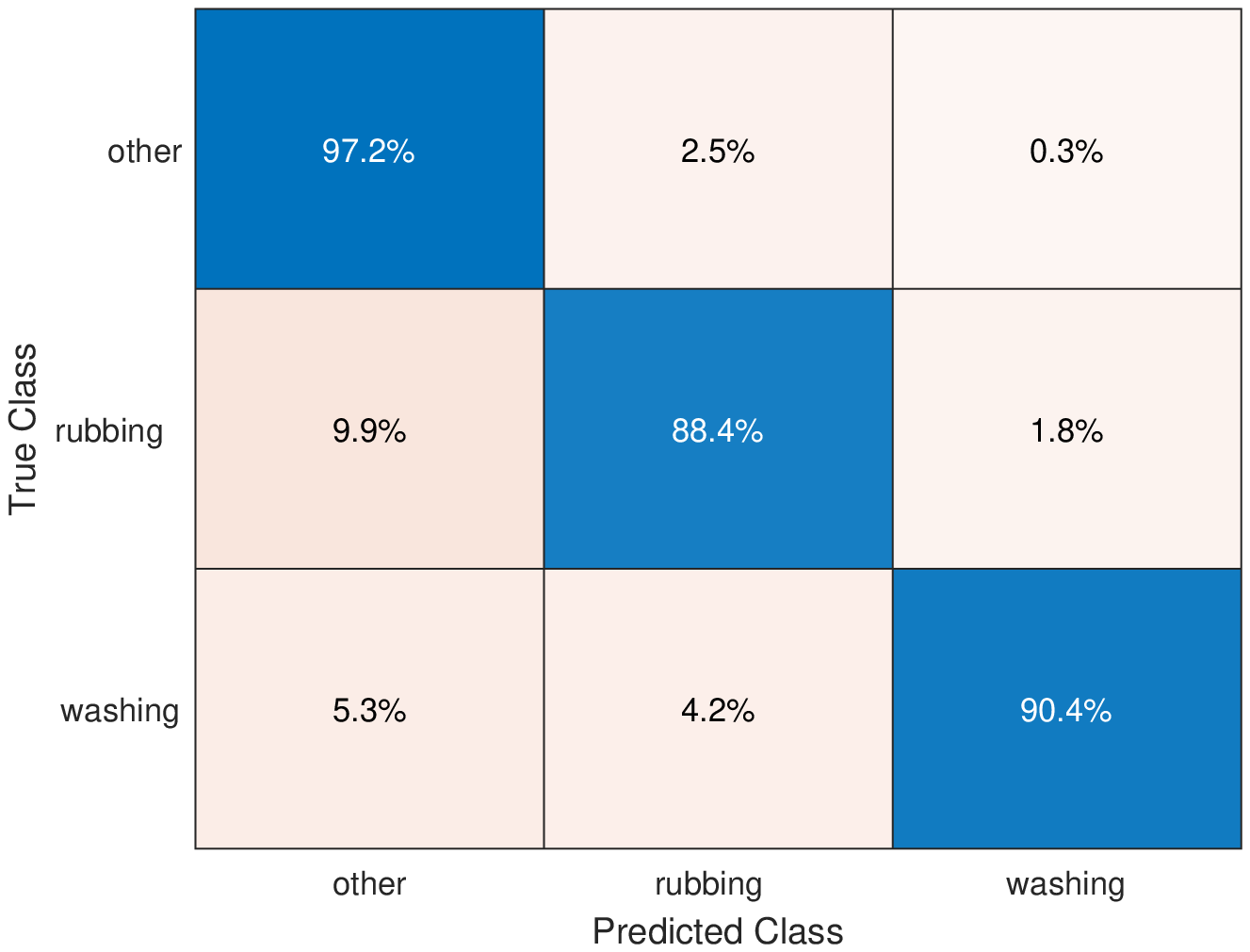}}
\subfloat[CNN]{\includegraphics[width=0.9\columnwidth]{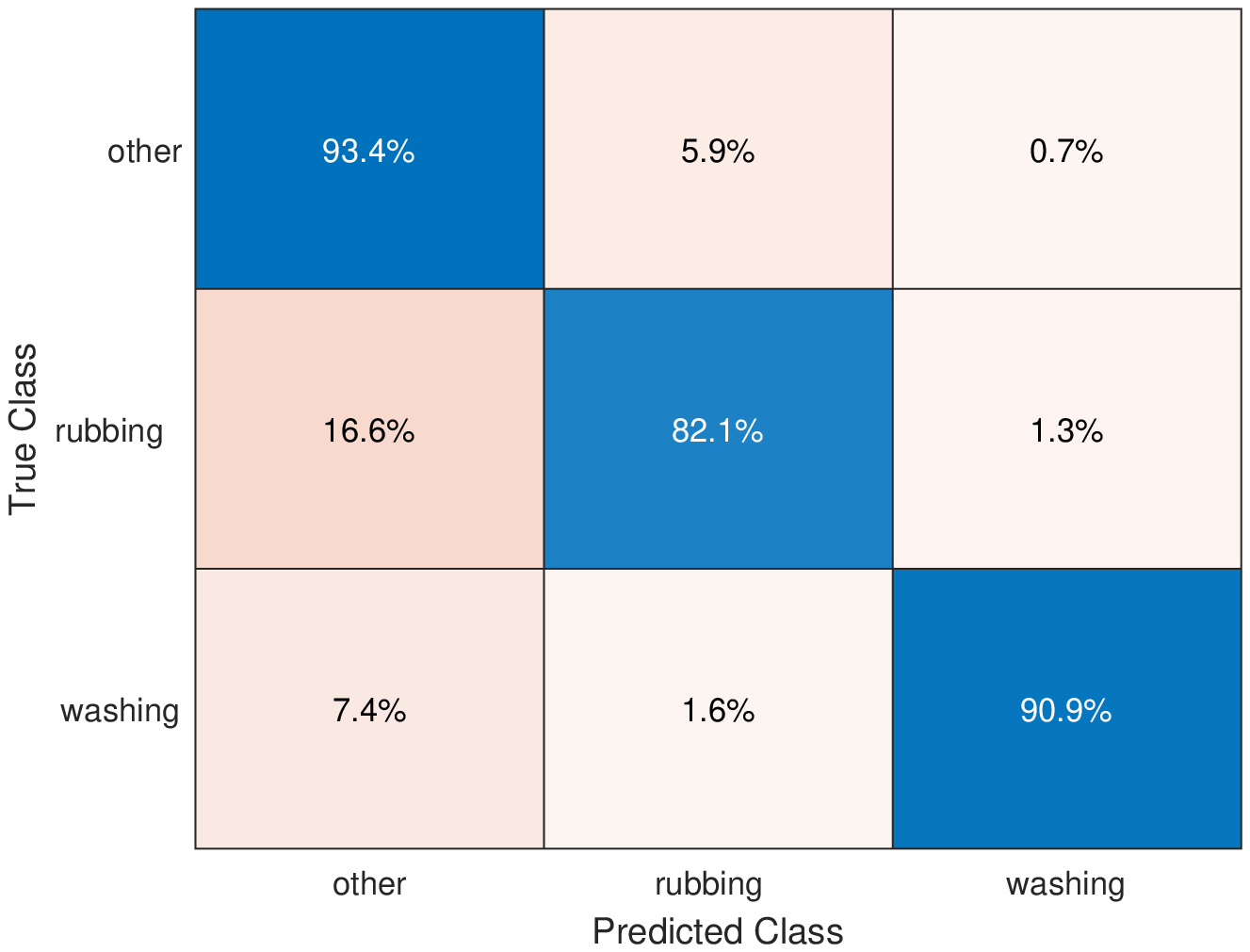}}
\caption{Average confusion matrices calculated on the 5-fold cross-validation test.}
\label{fig:confusion_best_activity_classification}
\end{figure*}

Figure~\ref{fig:confusion_best_activity_classification} reports the average confusion matrices calculated on top of the results obtained during the 5-fold cross-validation tests. All four models used show a great ability to correctly classify the {\it other} activity. For instance, the LSTM network reaches the higher value of about 97.2\%. Furthermore, also the {\it washing} activity has been correctly classified reaching the higher values of about 95\% using standard tools while deep learning models do not exceed 91\%. The {\it rubbing} activity, on the other hand, appears to be the most difficult to classify with a lower value of about 82\% obtained with the CNN model. Here, in fact, the {\it rubbing} class has been misclassified as {\it other} about 16\% of the time. This is probably due to the fact that handrubbing is a less dynamic activity with respect to hand washing which therefore produces fewer accelerations and rotations of the wrist. Furthermore, the use of running water for washing could introduce vibrations that are more easily identifiable by the classifiers.     

\subsubsection{Subject classification}
\label{sec:subject_classification_ad_hoc}

The second set of classification experiments has been carried out aimed at identifying the person washing or rubbing their hands instead of the performed activity. For this purpose, each sample related to the {\it other} activity has been removed from the database while {\it washing} and {\it rubbing} samples have been merged into a single class to which a label containing a unique person identifier has been added.  

\begin{table}[htbp]
  \caption{Best subject classification results obtained with the proposed models.}
  \centering
    \begin{tabular}{rcccc}
    \toprule
          & SVM & ERS-KNN & LSTM & CNN \\
    \midrule
    Accuracy   & $\mathbf{0.991}$ & $0.988$ & $0.966$ & $0.958$ \\
    Precision  & $\mathbf{0.990}$ & $0.985$ & $0.959$ & $0.945$ \\
    Recall     & $\mathbf{0.989}$ & $0.986$ & $0.956$ & $0.952$ \\
    F1-score   & $\mathbf{0.990}$ & $0.985$ & $0.957$ & $0.948$ \\
    \bottomrule
    \end{tabular}%
  \label{tab:subjects_classification}%
\end{table}%
  
Table~\ref{tab:subjects_classification} shows the best value of the classification metrics obtained with the four models. Also in this case, the best results have been obtained with the following size of the processing window: SVM=12s; ERS-KNN=8s; LSTM=2s; CNN=6s. In the case of standard tools, all the proposed features have been used.
As expected, recognizing the person who is washing/rubbing the hands is a much easier task thanks to the fact that the ad-hoc dataset contains data collected in an unstructured way where each subject is free to wash its hands as he/she is used to. Our results, with the higher accuracy of about $0.99$ obtained with the SVM classifier, in fact, suggest that the hand washing/rubbing activity can represent a kind of subject fingerprint. Another interesting result from this experiment concerns the fact that SVM and ERS-KNN seem to exceed the deep learning methods by almost $5$ percentage points.    

\subsection{Memory footprint and timing}

In order to indirectly evaluate the complexity of the proposed machine learning models, during each experiment, we measured the timing performances and the memory footprint of each classifier.
In particular, each classifier has been implemented on Matlab2021a\textsuperscript{\textregistered}~platform by means of Machine Learning tools and get executed on Intel\textsuperscript{\textregistered} Core i9 desktop PC equipped with 32GB of RAM and with an NVIDIA\textsuperscript{\textregistered} Quadro\textsuperscript{\textregistered} RTX\textsuperscript{\texttrademark} 4000 GPU. 
\begin{table}[htbp]
  \caption{Timing performance and memory footprint.}
  \centering
    \begin{tabular}{rrrrr}
    \toprule
          & SVM & ERS-KNN & LSTM & CNN \\
    \midrule
    Training time (s)     & $7.191$ & $\mathbf{5.803}$ & $472.069$ & $299.551$ \\
    Inference time (ms)   & $\mathbf{0.016}$ & $0.331$ & $1.749$ & $0.505$ \\
    Memory footprint (MB) & $\mathbf{4.168}$ & $41.996$ & $5.569$ & $6.389$ \\
    \bottomrule
    \end{tabular}%
  \label{tab:models_performances}%
\end{table}%
Both training and inference phases have been conducted by setting GPU as the execution environment in order to fully exploit the CUDA\textsuperscript{\textregistered} parallelism of the video adapter.  

Table~\ref{tab:models_performances} shows the timing performance and the memory footprint of each model with highlighted the best values. As expected the training phase of deep learning models is the most time expensive phase with a maximum value of about $470$ seconds measured in the LSTM network. On the other hand, the training phase of the ERS-KNN results to be the fastest with only about $5.8$ seconds. From the inference time point of view, the SVM significantly exceeds other models performances as it needs only about 160 \textmu seconds to infer a label. In particular, it results in an improvement in performance in the order of about 20x, 100x, and 40x with respect to, respectively, ERS-KNN, LSTM, and CNN. Notice that, the reported inference time for SVM and ERS-KNN also includes the feature extraction time.
If in the cloud-based approaches the computational cost of the training phase can be easily overcome by resorting to powerful GPU/TPU, the inference phase, on the other hand, should be done directly on the wearable device to explore several appealing benefits. For instance, executing the inference phase on the edge node avoids latency issues due to communications to and from the cloud, it enables higher levels of privacy and security by keeping most of the data on remote devices, and, finally, it can improve energy efficiency by trading off computation and communication energy requirements.
From this point of view, Table~\ref{tab:models_performances} shows that SVM, thanks to its low memory footprint and low inference time, can be the most suitable model for a real-time wearable application.

\subsection{Sensitivity to the window length}

\begin{figure*}[!h]
\centering
\subfloat[SVM]{\includegraphics[width=0.9\columnwidth]{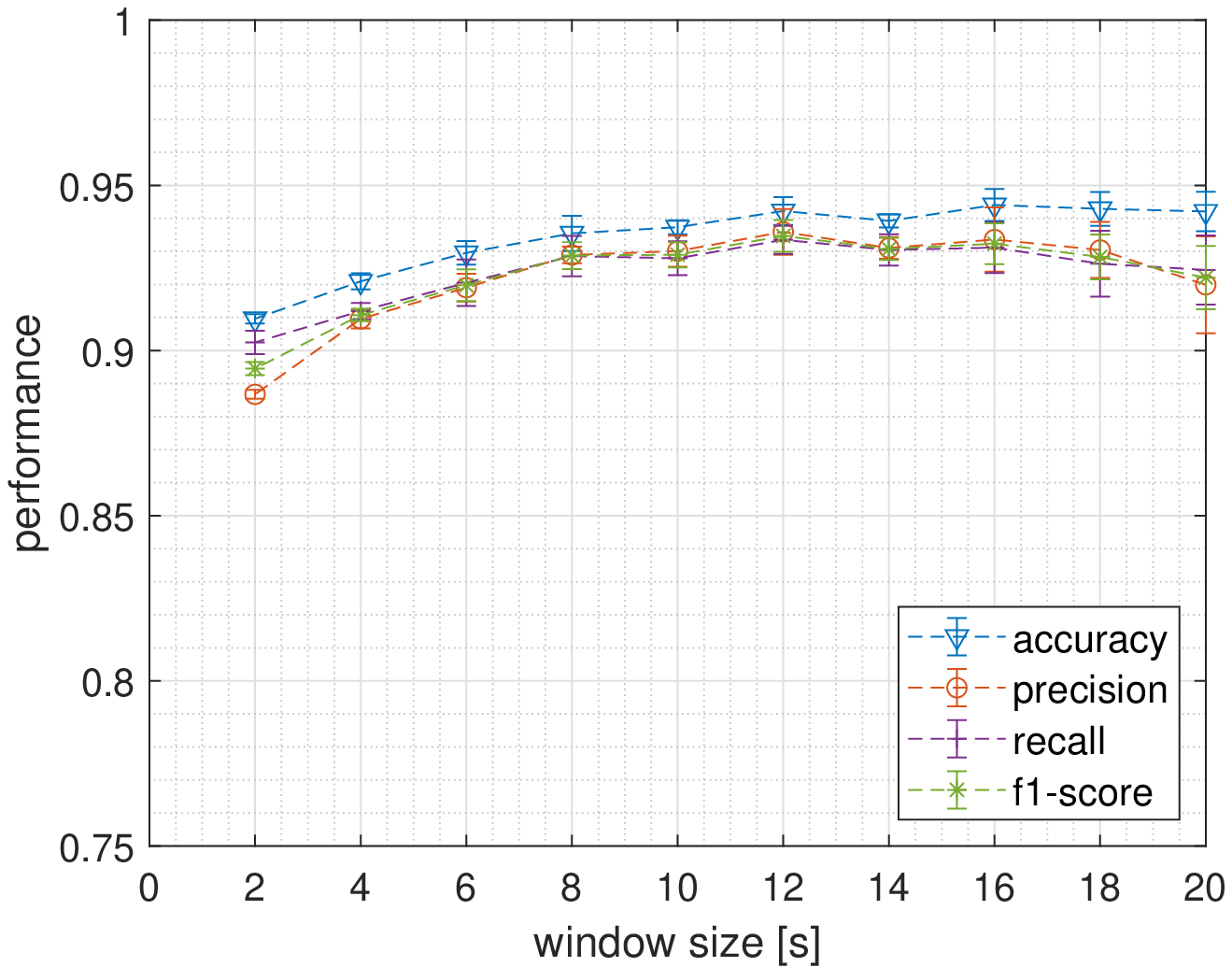}}
\subfloat[ERS-KNN]{\includegraphics[width=0.9\columnwidth]{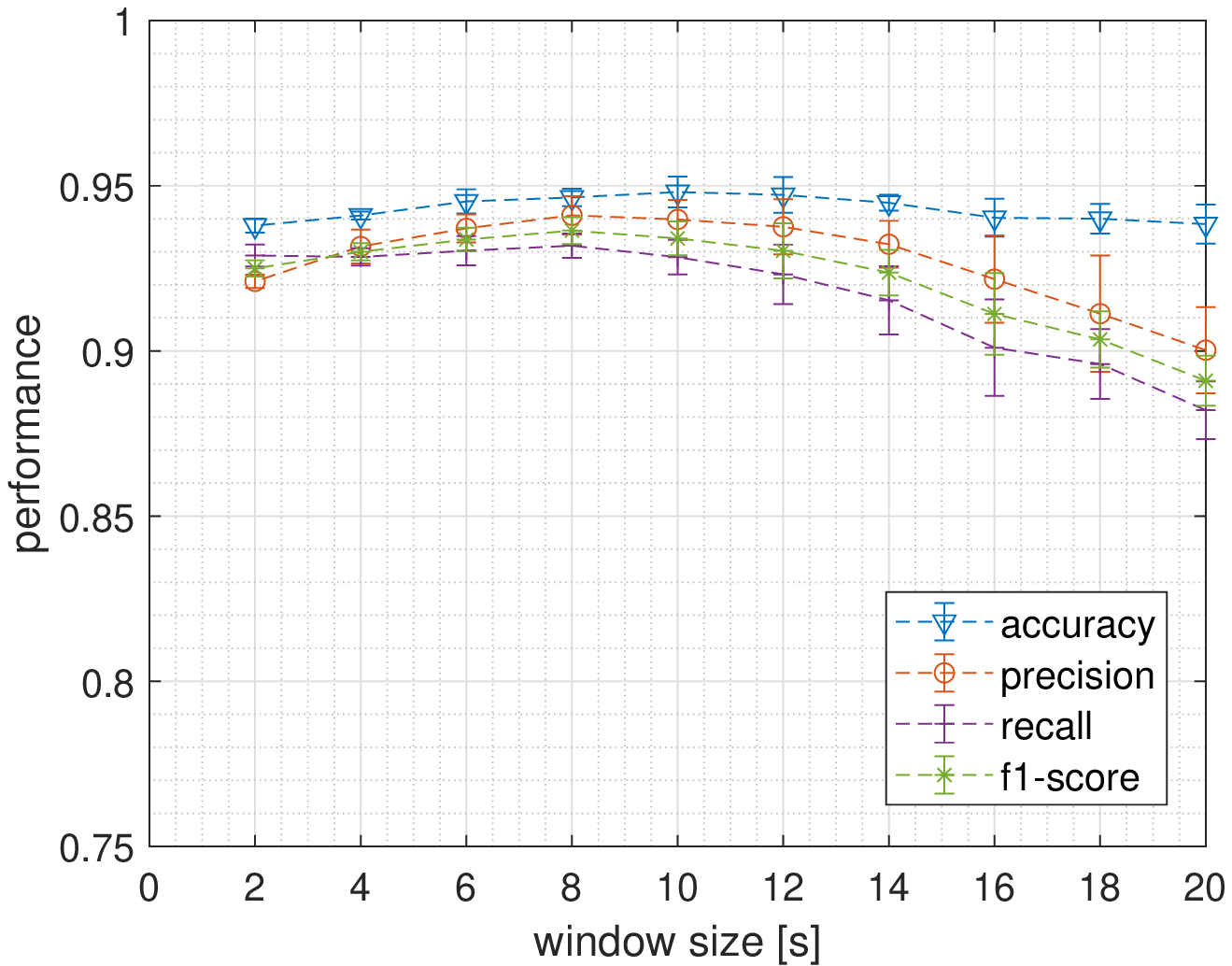}}\\
\subfloat[LSTM]{\includegraphics[width=0.9\columnwidth]{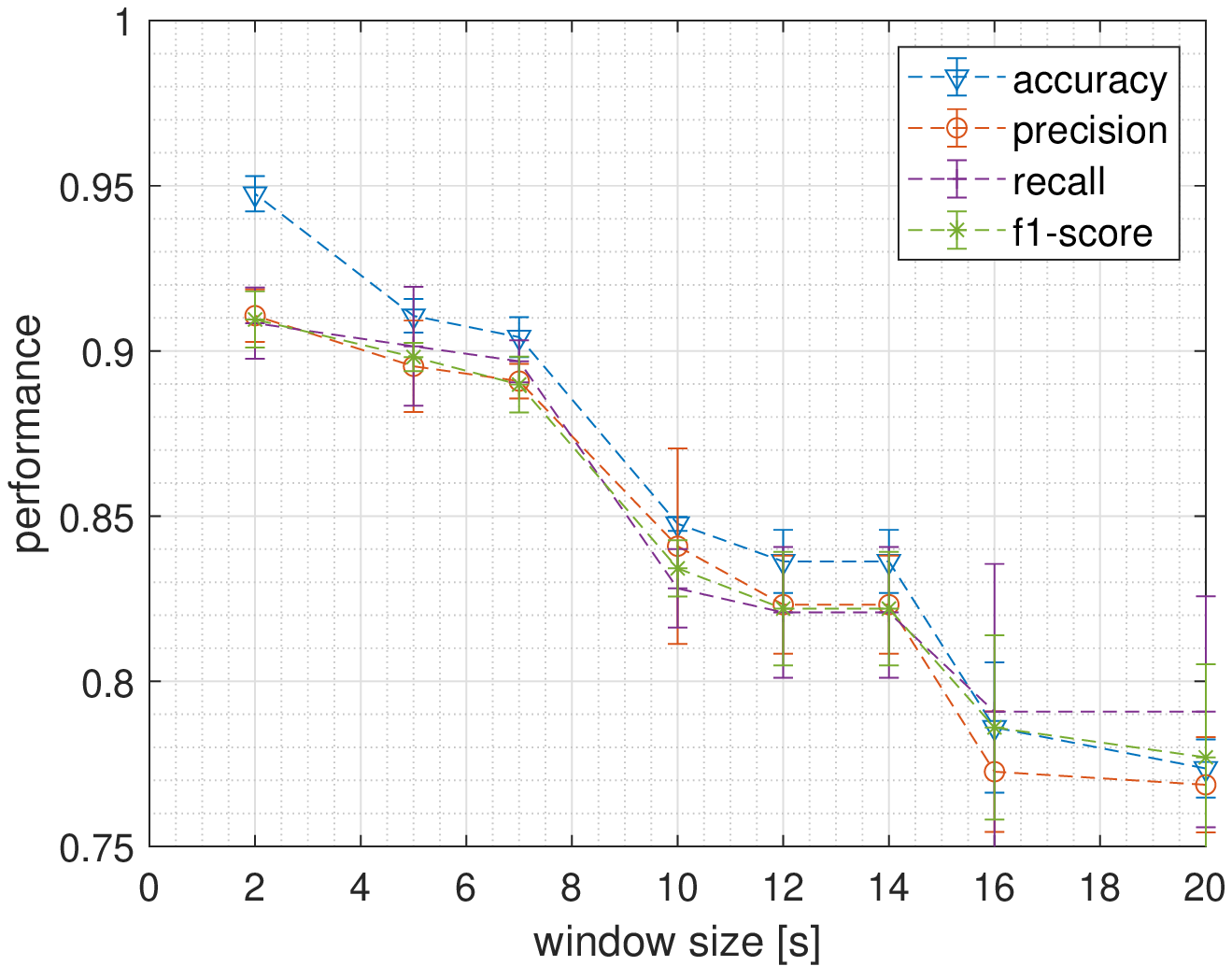}}
\subfloat[CNN]{\includegraphics[width=0.9\columnwidth]{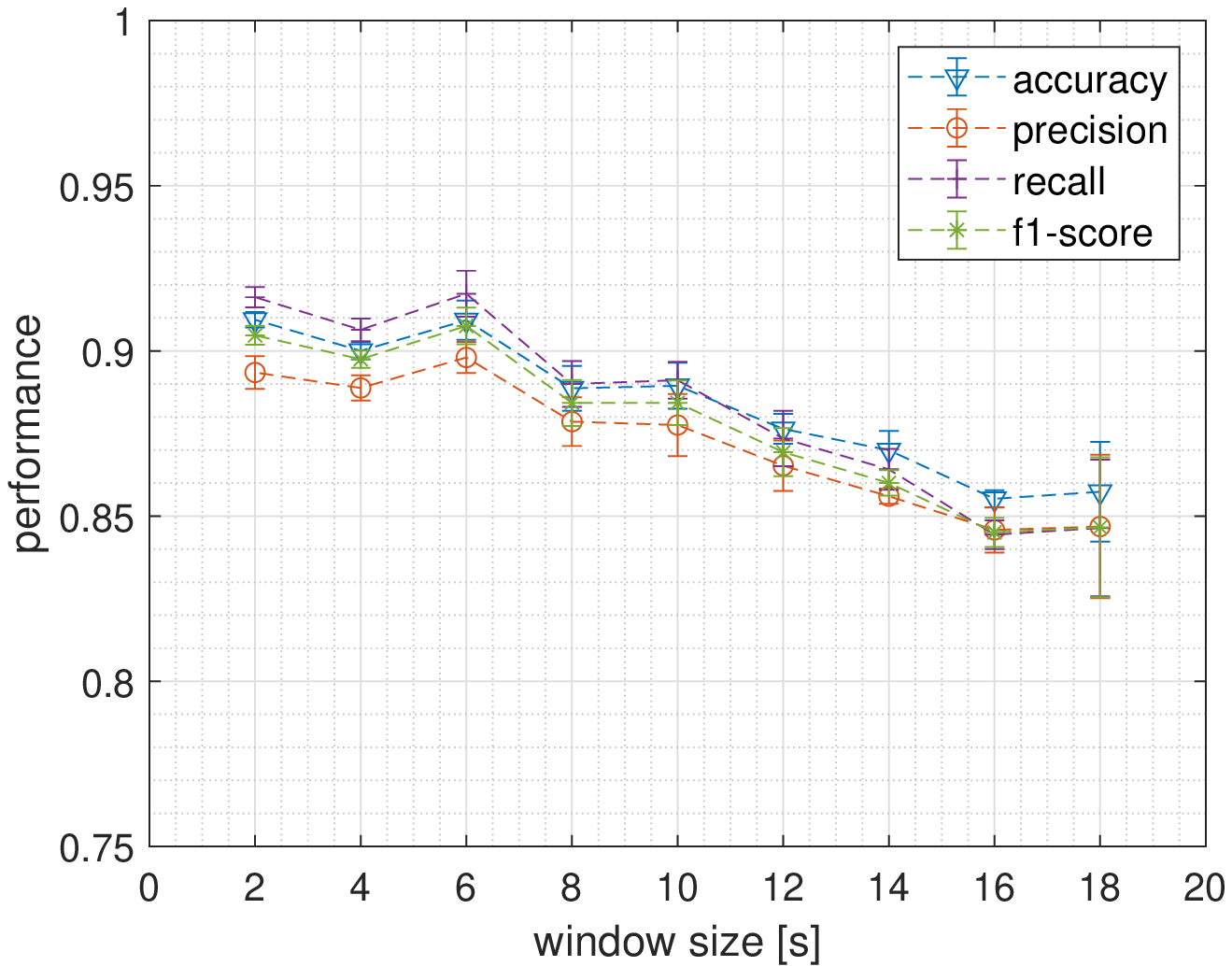}}

\caption{Performances of the proposed classifiers when varying the size of the processing window.}
\label{fig:window_size}
\end{figure*}


The size of the processing window influences the performance of the classification models in several ways. In this section, the results of the in-depth analysis of this dependence are reported. 
In particular, Figure~\ref{fig:window_size} plots the classification metrics obtained by the four classifiers when varying the size of the processing window. Notice that, in each experiment, we use a processing window with 75\% of overlap which leads to a total number of samples ranging from more than 200,000 to about 17,000 when the window increases from 2 to 20 seconds. 
Each point represents the average value together with the standard deviation calculated over a 5-fold cross-validation test.
Both SVM and ERS-KNN (Figure~\ref{fig:window_size}.(a) and Figure~\ref{fig:window_size}.(b)) show an almost flat trend of the measured accuracy even if at some point the other metrics (precision, recall, and f1-score) begin to deteriorate as the window size increases. In particular, the SVM classifier increases its performance until when using a window of about 12 seconds. Further increasing the size of the window leads to an average decrease in Precision, Recall, and F1-score and to more unstable results (higher standard deviations).

Similarly, the performances of ERS-KNN increase until a window size of about 8 seconds beyond which they markedly decrease together with results stability.  

An opposite trend is found, however, regarding the results obtained by the two deep learning classifiers (Figure~\ref{fig:window_size}.(c) and Figure~\ref{fig:window_size}.(d)). In this case, in fact, the four performance metrics show an almost monotonous decreasing trend for increasing values of the window size. Moreover, for the CNN classifier, we found a local maximum at a window size of about 6 seconds. 
 
\subsection{Non-parametric significance tests}
In order to statistically assess whether the accuracies of the four classification models are different, three variations of the McNemar test have been performed: i) asymptotic test; ii) exact-conditional test; iii) mid-{\it p}-value test~\cite{Fagerland2013}. These tests compare two classifiers by analyzing their predicted labels against the true labels and then detect whether the difference between the misclassification rates is statistically significant.

\begin{table*}[htbp]
\caption{Results of three variations of the McNemar test.}
  \centering
    \begin{tabular}{c|l|rr|rr|cr}
    \multicolumn{1}{r}{} & \multicolumn{1}{r}{} & \multicolumn{2}{c}{SVM} & \multicolumn{2}{c}{LSTM} & \multicolumn{2}{c}{CNN} \\
    \midrule
    \multicolumn{1}{r}{} &       & \multicolumn{1}{c}{h} & \multicolumn{1}{c|}{p} & \multicolumn{1}{c}{h} & \multicolumn{1}{c|}{p} & h     & \multicolumn{1}{c}{p} \\
    \midrule
    \multirow{3}[2]{*}{ERS-KNN} & asymptotic & \multicolumn{1}{c}{\textbf{false}} & 0.65  & \multicolumn{1}{c}{true} & \multicolumn{1}{l|}{$4.14\times10^{-25}$} & true  & \multicolumn{1}{l}{$2.04\times10^{-21}$} \\
          & mid-p & \multicolumn{1}{c}{\textbf{false}} & 0.65  & \multicolumn{1}{c}{true} & \multicolumn{1}{l|}{$1.33\times10^{-25}$} & true  & \multicolumn{1}{l}{$3.12\times10^{-27}$} \\
          & exact & \multicolumn{1}{c}{\textbf{false}} & 0.69  & \multicolumn{1}{c}{true} & \multicolumn{1}{l|}{$1.82\times10^{-32}$} & true  & \multicolumn{1}{l}{$1.12\times10^{-22}$} \\
    \midrule
    \multirow{3}[2]{*}{SVM} & asymptotic &       &       & \multicolumn{1}{c}{true} & $3.94\times10^{-05}$  & true & $4.53\times10^{-22}$ \\
          & mid-p &       &       & \multicolumn{1}{c}{true} & $3.87\times10^{-05}$  & true & $1.27\times10^{-18}$ \\
          & exact &       &       & \multicolumn{1}{c}{true} & $4.46\times10^{-05}$  & true & $3.43\times10^{-22}$ \\
    \midrule
    \multirow{3}[2]{*}{LSTM} & asymptotic &       &       &       &       & true &  $1.23\times10^{-25}$ \\
          & mid-p &       &       &       &       & true &  $5.13\times10^{-25}$ \\
          & exact &       &       &       &       & true &  $2.26\times10^{-24}$ \\
    \bottomrule
    \end{tabular}%
  \label{tab:significance}%

\end{table*}%

Table~\ref{tab:significance} reports the results of the three tests conducted for each couple of classifiers. The possible combinations of the four classifiers produce a symmetric matrix which, for readability, is reported only in its upper part. In the same way, the comparison between the classifiers and itself has not been carried out (the matrix diagonal) for the obviousness of the results. 

For each test, the logical value {\it h} is reported which represents the test decision when testing the null hypothesis that the two classifiers have equal accuracy for predicting the true class. So, a false value indicates that the null hypothesis is not rejected with a confidence level of 95\% ($p < 0.05$). 
Moreover, also the {\it p} value is reported which represents how strong is the evidence to reject or not the null hypothesis. 

For instance, when comparing ERS-KNN with SVM, the three variants of the McNemar test agree not to reject the null hypothesis while, for both ERS-KNN Vs LSTM and ERS-KNN Vs CNN, the null hypothesis needs to be rejected so that the accuracies of the two classification models can not be considered equivalent. Furthermore, in these cases, the {\it p}-value for each test is close to zero, which indicates strong evidence to reject the null hypothesis that the two classifiers have equal predictive accuracies. 

Also the comparison between SVM and LSTM, SVM and CNN, and LSTM and CNN, lead to strong rejections of the null hypothesis with {\it p}-values very close to zero so that, in this scenario, only standard machine learning tools, namely SVM and ERS-KNN, can be considered statistically equivalent from the classification performance point of view.
      
\subsection{Feature selection results}


In order to evaluate the relative influence of the proposed features on the classification performances, we use the forward feature selection method~\cite{liu2005}. 
Forward feature selection is based on an objective function (e.g. the accuracy) which is used as a criterion to evaluate the impact of adding a feature from a candidate subset, starting from an empty set until adding other features doesn't induce any improvement in the objective function.
We applied this strategy to highlight how the proposed features contribute to the overall performance of the two standard classifiers. In particular, each group of features, namely {\it Base (B)}, {\it Hjorth (H)}, and {\it Shape (S)} has been treated as an atomic unit that can be added or removed as a whole. First of all, we tested each classifier using only one of the three groups, and then we added the other groups to explore all possible combinations. 

\begin{table*}[htbp]
    \caption{Results of the forward feature selection method applied to SVM classifier}
  \centering
    \begin{tabular}{lcccc}
    \toprule
          & Accuracy  & Precision  & Recall  & F1-s \\
    \midrule
    B        & $0.901 \pm 0.007$ & $0.892 \pm 0.008$ & $0.891 \pm 0.014$ & $0.891 \pm 0.006$ \\
    H        & $0.923 \pm 0.003$ & $0.912 \pm 0.005$ & $0.913 \pm 0.006$ & $0.891 \pm 0.005$ \\
    S        & $0.339 \pm 0.098$ & $0.500 \pm 0.069$ & $0.480 \pm 0.067$ & $0.490 \pm 0.067$ \\
    B+H      & $0.941 \pm 0.006$ & $0.932 \pm 0.008$ & $0.932 \pm 0.006$ & $0.932 \pm 0.006$ \\
    B+S      & $0.916 \pm 0.004$ & $0.910 \pm 0.006$ & $0.906 \pm 0.010$ & $0.908 \pm 0.006$ \\
    H+S      & $0.926 \pm 0.011$ & $0.917 \pm 0.009$ & $0.921 \pm 0.067$ & $0.919 \pm 0.010$ \\
    B+H+S    & $\mathbf{0.942 \pm 0.008}$ & $\mathbf{0.936 \pm 0.008}$ & $\mathbf{0.934 \pm 0.006}$ & $\mathbf{0.935 \pm 0.007}$ \\
    \bottomrule
    \end{tabular}%
  \label{tab:feature_selection_SVM}%
\end{table*}%

Table~\ref{tab:feature_selection_SVM} shows the activities classification performances, together with its standard deviations, of the SVM when varying the adopted features. For each performance metric, the maximum value achieved has been highlighted in bold. All metrics showed a monotone increasing trend when consecutively adding one of three groups of features reaching the higher performances when all the proposed features are used together ({\it Base+Hjorth+Shape}).
This suggests that all features provide original information content useful for the classification process. 
 Furthermore, the {\it Hjorth} group seems to contain the most informative group of features producing the highest classification performance with respect to the other groups when tested alone.  

The same experiment conducted with the ESR-KNN classifier produces comparable results, reported in Table~\ref{tab:feature_selection_ESKNN}, with the only difference that, in this case, the measured performances are slightly higher.  

\begin{table*}[htbp]
  \caption{Results of the forward feature selection method applied to ERS-KNN classifier}
  \centering
    \begin{tabular}{lcccc}
    \toprule
          & Accuracy  & Precision  & Recall & F1-score \\
    \midrule
    B             & $0.904 \pm 0.007$ & $0.899 \pm 0.003$ & $0.873 \pm 0.009$ & $0.886 \pm 0.002$ \\
    H            & $0.908 \pm 0.004$ & $0.908 \pm 0.005$ & $0.889 \pm 0.006$ & $0.886 \pm 0.005$ \\
    S            & $0.719 \pm 0.007$ & $0.683 \pm 0.010$ & $0.674 \pm 0.009$ & $0.678 \pm 0.010$ \\
    B+H       & $0.944 \pm 0.002$ & $0.937 \pm 0.003$ & $0.929 \pm 0.002$ & $0.933 \pm 0.002$ \\
    B+S       & $0.908 \pm 0.006$ & $0.909 \pm 0.006$ & $0.886 \pm 0.009$ & $0.897 \pm 0.007$ \\
    H+S      & $0.922 \pm 0.004$ & $0.920 \pm 0.006$ & $0.902 \pm 0.009$ & $0.911 \pm 0.006$ \\
    B+H+S & $\mathbf{0.946 \pm 0.003}$ & $\mathbf{0.941 \pm 0.003}$ & $\mathbf{0.932 \pm 0.003}$ & $\mathbf{0.936 \pm 0.003}$ \\
    \bottomrule
    \end{tabular}%
  \label{tab:feature_selection_ESKNN}%
\end{table*}%

Notice that, for both classifiers, these results suggest that a good trade-off between classification performances and real-time computation complexity can be represented by the design choice of calculating only {\it Base+Hjorth} giving up only about 0.2\% of classification performance decrease. Moreover, if for the purpose of a particular real-time application, an accuracy of about 92\% could be considered acceptable, it even would be sufficient to calculate the Hjorth features, saving many computational resources and power.

\subsection{Classification results on DLA dataset}
\label{sec:resultDLA}

The models described in the previous sections have been trained and tested on the publicly available DLA dataset. 
A first set of experiments has been made to classify the following ten activities: {\it handwriting, handwashing, facewashing, teethbrush, sweeping, vacuuming, eating, dusting, rubbing, other}, and, a second set has been made in order to recognize the 8 subjects of the dataset.

\subsubsection{Human activity classification}

\begin{table}[htbp]
  \caption{Best activity classification results obtained with the proposed models on DLA dataset.}
  \centering
    \begin{tabular}{rcccc}
    \toprule
          & SVM & ERS-KNN & LSTM & CNN \\
    \midrule
    Accuracy    & $0.891$ & $0.914$ & $0.866$ & $\mathbf{0.923}$ \\
    Precision   & $0.907$ & $0.931$ & $0.882$ & $\mathbf{0.939}$ \\
    Recall      & $0.867$ & $0.914$ & $0.874$ & $\mathbf{0.925}$ \\
    F1-score    & $0.886$ & $0.922$ & $0.878$ & $\mathbf{0.932}$ \\
    \bottomrule
    \end{tabular}%
  \label{tab:activities_classificationDLA}%
\end{table}%

Table~\ref{tab:activities_classificationDLA} reports the classification results obtained with the models configuration and parameters used in section~\ref{sec:activity_classification_ad_hoc}. 
Each value is reported as the average value calculated during the 5-fold cross-validation test. For each metric, the highest value obtained ever is highlighted in bold.

\begin{figure*}[!h]
\centering
\subfloat[SVM]{\includegraphics[width=0.9\columnwidth]{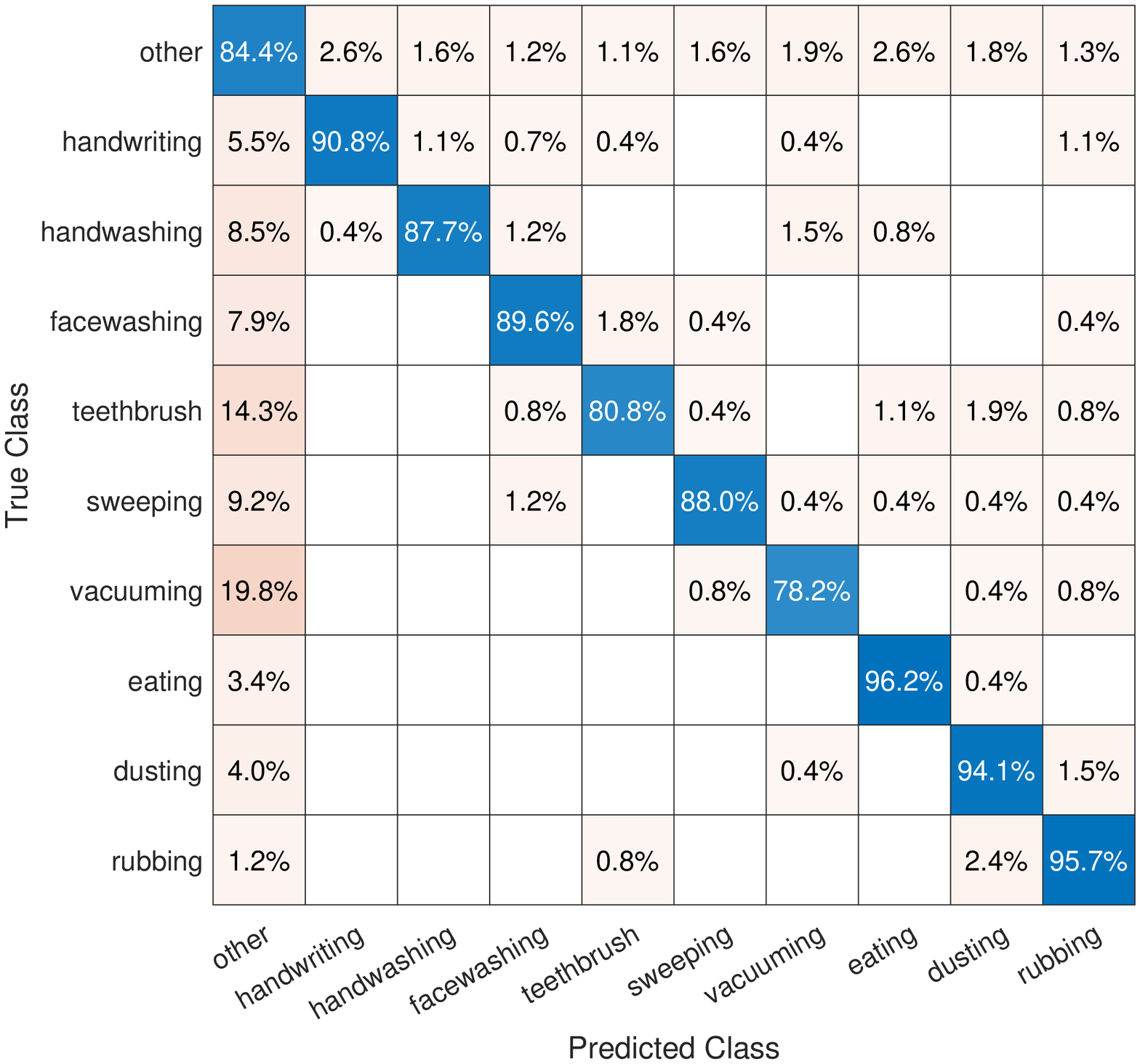}}
\subfloat[ERS-KNN]{\includegraphics[width=0.9\columnwidth]{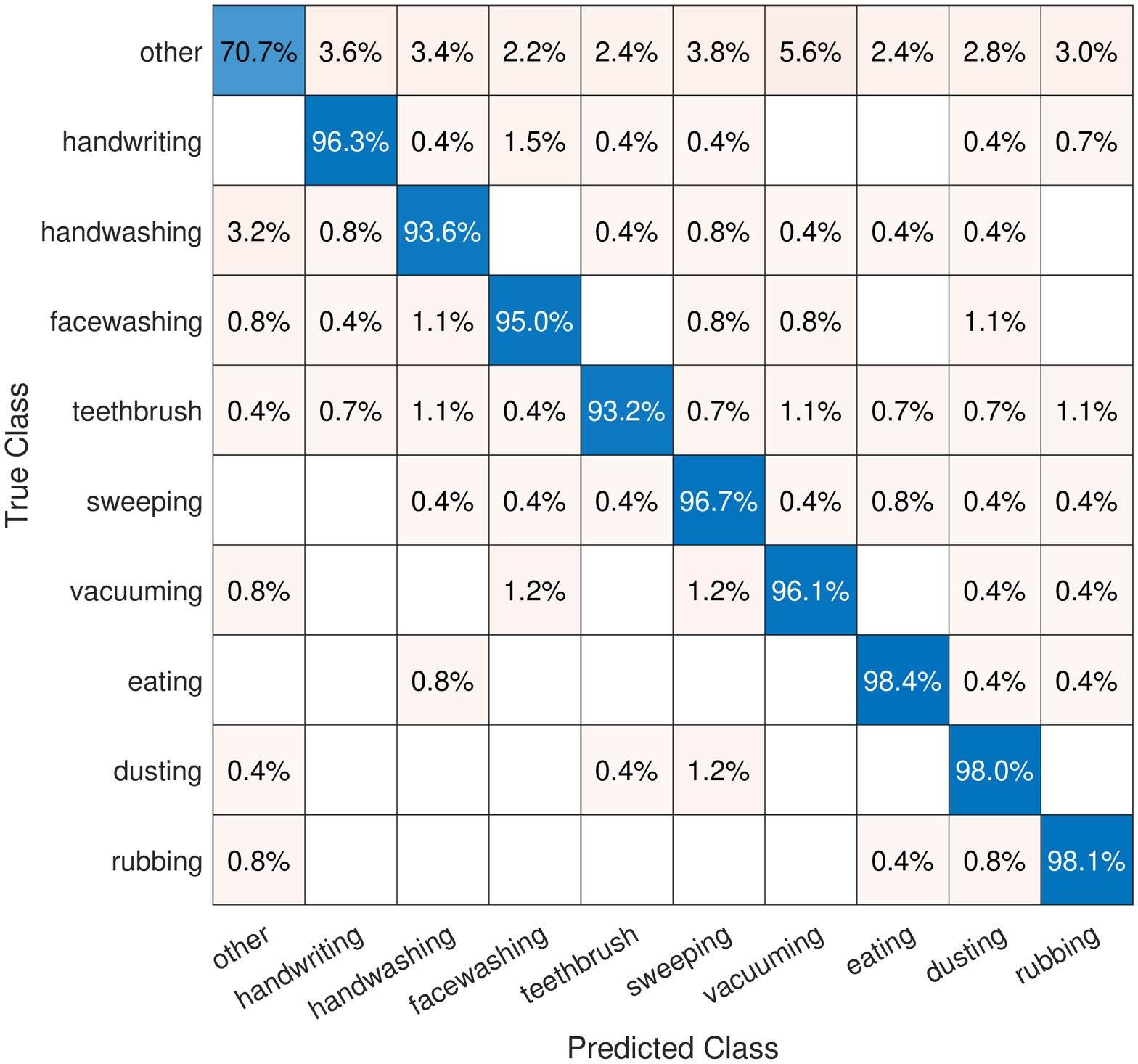}}\\
\subfloat[LSTM]{\includegraphics[width=0.9\columnwidth]{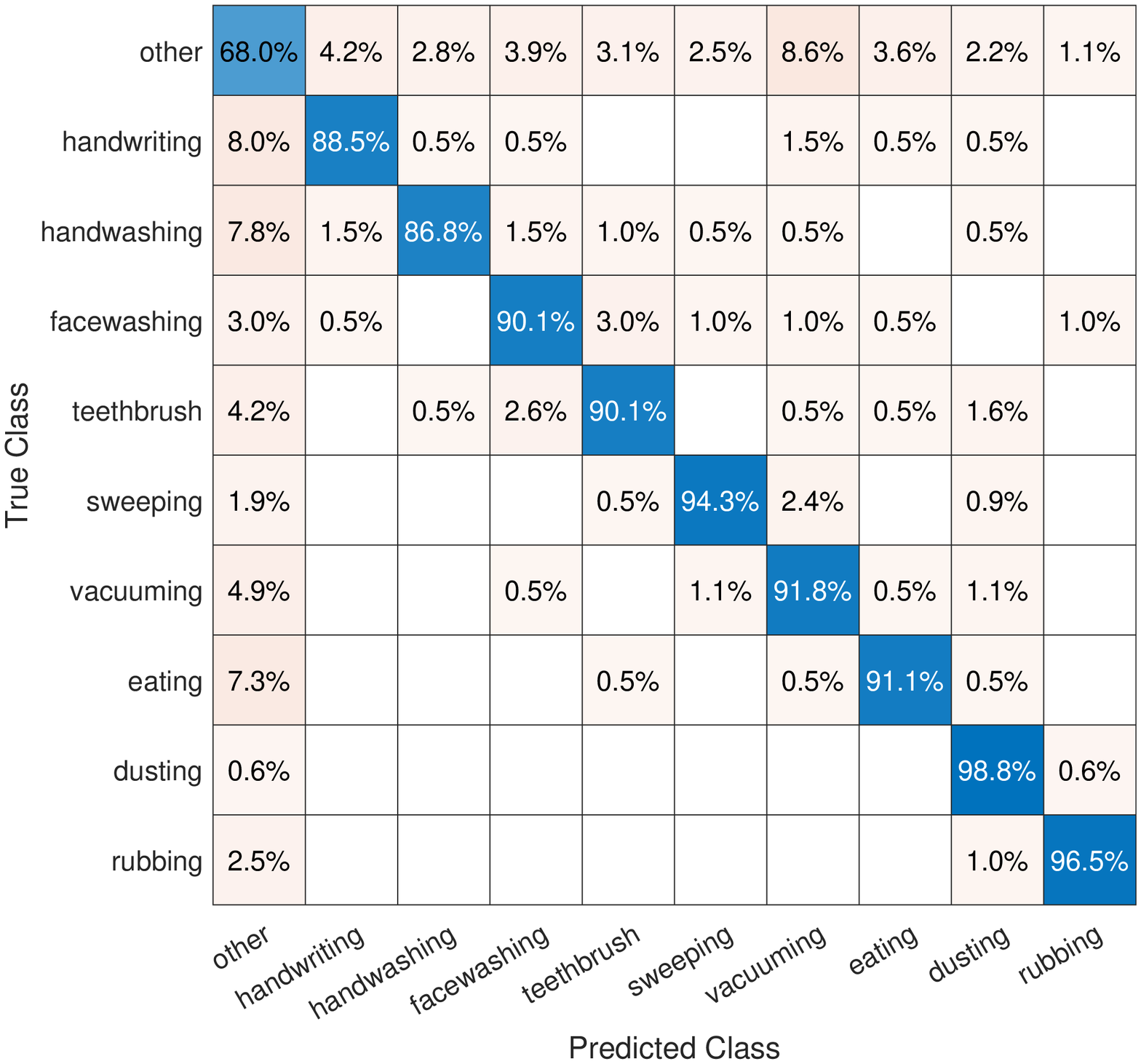}}
\subfloat[CNN]{\includegraphics[width=0.9\columnwidth]{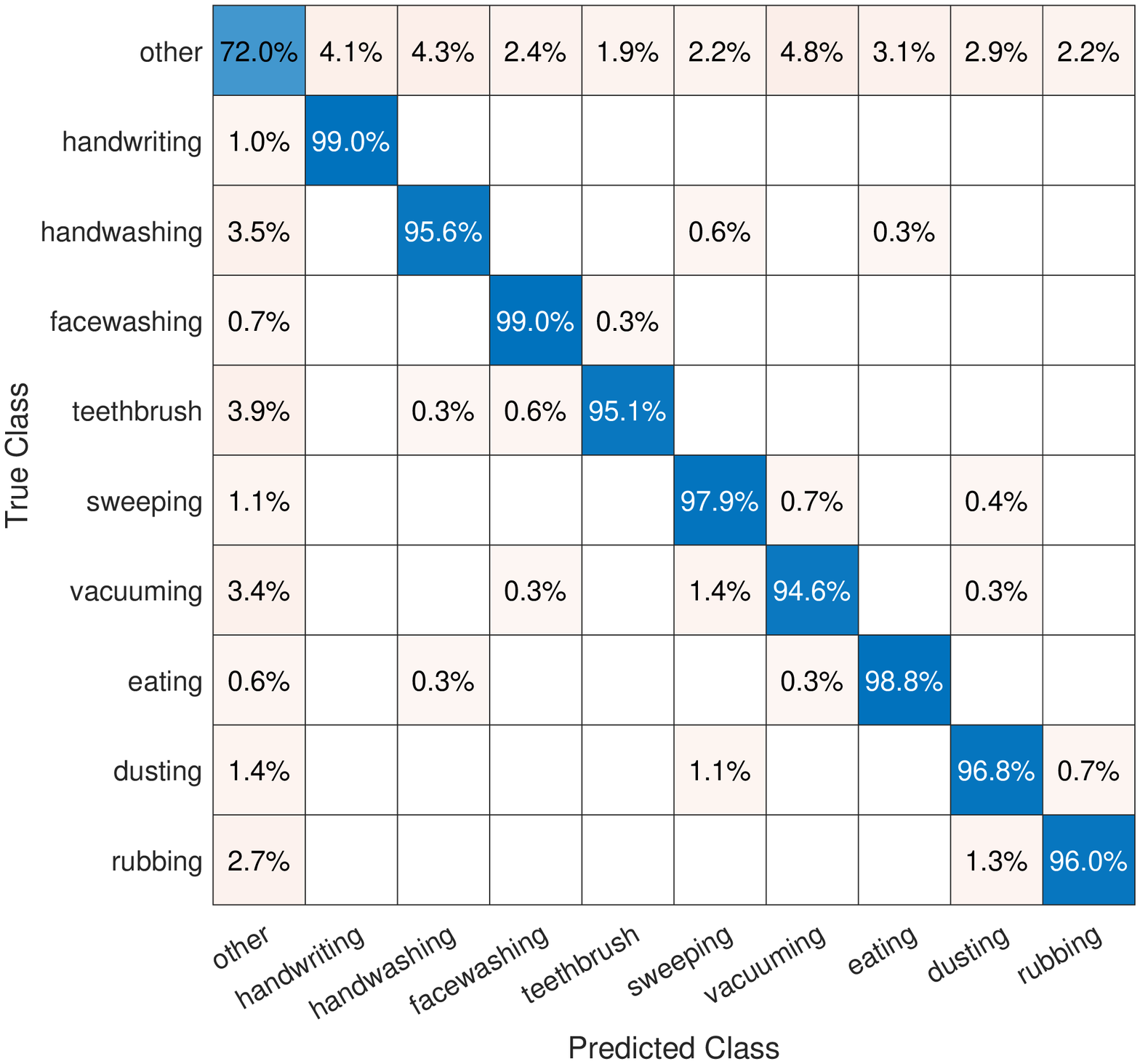}}
\caption{Average confusion matrices calculated on the 5-fold cross-validation test on DLA dataset.}
\label{fig:confusion_best_activity_classification_DLA}
\end{figure*}

In this experiment, the CNN outperforms other models in each of the calculated metrics. 
Regarding the ERS-KNN performance, the gap is about 1\% or 2\% (depending on the considered metric) while, it increases up to 6\% for some metrics calculated for SVM and LSTM models
Despite this, the performance of ERS-KNN and CNN remains well above 91\% in each computed metric demonstrating a high capability in classifying wrist human activities using signals from wearable IMU devices.

Compared to the performances of the experiment conducted on the ad-hoc dataset, the average values suffered a decrease of about 1\% or 2\% for ERS-KNN and CNN and of about 7\% and 8\% for SVM and LSTM. A generalized decrease in classification performance is certainly due to the greater number of classes used in this second dataset, which can lead to a greater dispersion of the results. Furthermore, the lack of gyroscopic signals may have had a significant impact on classification capabilities, especially for SVM and LSTM which show the lowest values.

Figure~\ref{fig:confusion_best_activity_classification_DLA} reports the average confusion matrices calculated on top of the results obtained during the 5-fold cross-validation tests. ERS-KNN and CNN 
demonstrate a great ability to correctly classify the activity of interest (i.e. the {\it handwashing}). For instance, CNN correctly recognizes it over 95\% of the time while the ERS-KNN about 94\% of the time. On the other hand, for the SVM and LSTM models, this rate decreases up to 88\% and 87\% respectively. 
Moreover, the remaining wrong classifications of the {\it handwashing} activity are due overall to its misclassification as {\it other} and they are not due to another wrist activity in particular, effectively representing false negatives.

\subsubsection{Subject classification}

As for the subject classification on ad-hoc dataset experiments, each sample related to the {\it other} activity has been removed, while the remaining classes have been merged into a single class to which a label containing a unique person identifier has been added.

\begin{table}[htbp]
  \caption{Best subject classification results obtained with the proposed models on DLA dataset.}
  \centering
    \begin{tabular}{rcccc}
    \toprule
          & SVM & ERS-KNN & LSTM & CNN \\
    \midrule
    Accuracy   & $0.946$ & $\mathbf{0.959}$ & $0.902$ & $0.941$ \\
    Precision  & $0.949$ & $\mathbf{0.957}$ & $0.902$ & $0.940$ \\
    Recall     & $0.942$ & $\mathbf{0.949}$ & $0.903$ & $0.943$ \\
    F1-score   & $0.946$ & $\mathbf{0.959}$ & $0.903$ & $0.939$ \\
    \bottomrule
    \end{tabular}%
  \label{tab:subjects_classification_DLA}%
\end{table}%

Table~\ref{tab:subjects_classification_DLA} shows the value of the classification metrics obtained with the four models. Also for the DLA dataset, recognizing the person who is performing the wrist-based action is an easier task with respect to recognizing the activity. Moreover, also in this experiment, the standard classifiers (i.e. SVM and ERS-KNN) overcome the performance of the deep learning methods even if, with respect to CNN, the gap is appreciably reduced.

In general, despite this experiment is showing a high average classification accuracy of the proposed models (greater than 95\%), it also highlights an average reduction of performances while switching from the ad-hoc dataset to the DLA. Also in this case, in all probability, the loss of performance can be attributed to the higher number of subjects to be recognized and to the lack of the gyroscopic signals.

\section{Conclusions}
\label{sec:conclusions}
Hands hygiene is extremely important in breaking the chain of pathogens transmission by contact. In fact, contaminated hands are a privileged way to get to the mucous membrane of the mouth, nose, or eyes. Also regarding the COVID-19, it is estimated that a non-negligible part of infections occurs due to contact, through our hands, with contaminated surfaces.

In this work, we proposed and evaluated four classification machine learning models to distinguish the unstructured handwashing/handrubbing gestures from the rest of the daily activities starting from commonly used wearable devices. The proposed models lay the foundations for the creation of a system that provides users with automatic and continuous indirect monitoring of hands hygiene in an attempt to reduce the contact transmission of pathogens including SARS-CoV-2 coronavirus.

The experimental results, obtained over two different datasets containing in total more than 50 hours of recording of daily activities registration performed by 12 different subjects, show that both standard and deep learning techniques can be considered a viable solution to the classification problem reaching, respectively, an average accuracy of about 94\% and 95\%. Furthermore, considering the design of a low-cost low-power wearable device as a possible target of this work, our results show that SVM, thanks to its low memory footprint and low inference time, could allow executing the inference phase on the edge without recurring to the cloud infrastructure. This could, in principle, avoid latency and energy issues due to communications to and from the cloud, leading to higher levels of privacy and security by keeping most of the data on remote devices. Finally, as a last noteworthy result, we also show that by making use of the gyroscopic signals coming from the IMU devices it is possible to increase the recognition capability of standard learning tools, such as SVM and ERS-KNN, achieving the performances of the most complex model without impairing the computational cost.

\bibliographystyle{unsrt}
\bibliography{references}

\begin{thebibliography}{10}

\bibitem{santarpia2020}
Joshua~L Santarpia, Danielle~N Rivera, Vicki~L Herrera, M~Jane Morwitzer,
  Hannah~M Creager, George~W Santarpia, Kevin~K Crown, David~M Brett-Major,
  Elizabeth~R Schnaubelt, M~Jana Broadhurst, et~al.
\newblock Aerosol and surface contamination of sars-cov-2 observed in
  quarantine and isolation care.
\newblock {\em Scientific reports}, 10(1):1--8, 2020.

\bibitem{Wang2020}
Chaofan Wang, Zhanna Sarsenbayeva, Xiuge Chen, Tilman Dingler, Jorge Goncalves,
  and Vassilis Kostakos.
\newblock {Accurate Measurement of Handwash Quality Using Sensor Armbands:
  Instrument Validation Study}.
\newblock {\em JMIR mHealth and uHealth}, 8(3):e17001, mar 2020.

\bibitem{zhang2013}
Mi~Zhang and Alexander~A Sawchuk.
\newblock Human daily activity recognition with sparse representation using
  wearable sensors.
\newblock {\em IEEE journal of Biomedical and Health Informatics},
  17(3):553--560, 2013.

\bibitem{sztyler2016}
Timo Sztyler and Heiner Stuckenschmidt.
\newblock On-body localization of wearable devices: An investigation of
  position-aware activity recognition.
\newblock In {\em 2016 IEEE International Conference on Pervasive Computing and
  Communications (PerCom)}, pages 1--9. IEEE, 2016.

\bibitem{sztyler2017}
Timo Sztyler, Heiner Stuckenschmidt, and Wolfgang Petrich.
\newblock Position-aware activity recognition with wearable devices.
\newblock {\em Pervasive and mobile computing}, 38:281--295, 2017.

\bibitem{bhat2018}
Ganapati Bhat, Ranadeep Deb, Vatika~Vardhan Chaurasia, Holly Shill, and Umit~Y
  Ogras.
\newblock Online human activity recognition using low-power wearable devices.
\newblock In {\em 2018 IEEE/ACM International Conference on Computer-Aided
  Design (ICCAD)}, pages 1--8. IEEE, 2018.

\bibitem{koping2018}
Lukas K{\"o}ping, Kimiaki Shirahama, and Marcin Grzegorzek.
\newblock A general framework for sensor-based human activity recognition.
\newblock {\em Computers in biology and medicine}, 95:248--260, 2018.

\bibitem{lattanzi2022}
Emanuele Lattanzi, Matteo Donati, and Valerio Freschi.
\newblock Exploring artificial neural networks efficiency in tiny wearable
  devices for human activity recognition.
\newblock {\em Sensors}, 22(7):2637, 2022.

\bibitem{cheng2010}
Jingyuan Cheng, Oliver Amft, and Paul Lukowicz.
\newblock Active capacitive sensing: Exploring a new wearable sensing modality
  for activity recognition.
\newblock In {\em International conference on pervasive computing}, pages
  319--336. Springer, 2010.

\bibitem{singh2017}
Deepika Singh, Erinc Merdivan, Sten Hanke, Johannes Kropf, Matthieu Geist, and
  Andreas Holzinger.
\newblock Convolutional and recurrent neural networks for activity recognition
  in smart environment.
\newblock In {\em Towards integrative machine learning and knowledge
  extraction}, pages 194--205. Springer, 2017.

\bibitem{hassan2018}
Mohammed~Mehedi Hassan, Md~Zia Uddin, Amr Mohamed, and Ahmad Almogren.
\newblock A robust human activity recognition system using smartphone sensors
  and deep learning.
\newblock {\em Future Generation Computer Systems}, 81:307--313, 2018.

\bibitem{hou2020}
C.~{Hou}.
\newblock A study on imu-based human activity recognition using deep learning
  and traditional machine learning.
\newblock In {\em 2020 5th International Conference on Computer and
  Communication Systems (ICCCS)}, pages 225--234, 2020.

\bibitem{muller2021}
Heimo Muller, Michaela~Theresia Mayrhofer, Evert-Ben Van~Veen, and Andreas
  Holzinger.
\newblock The ten commandments of ethical medical ai.
\newblock {\em Computer}, 54(07):119--123, 2021.

\bibitem{surewash2021}
Glanta.
\newblock Surewash: The science of hand hygiene, 2021.
\newblock Online accessed 26 July 2021.

\bibitem{jimaging2020}
Chengzhang Zhong, Amy~R. Reibman, Hansel~A. Mina, and Amanda~J. Deering.
\newblock Multi-view hand-hygiene recognition for food safety.
\newblock {\em Journal of Imaging}, 6(11), 2020.

\bibitem{yue2021}
Zhentian Yue, Fuyong Wang, and Zhongxin Liu.
\newblock An intelligent hand-washing monitoring platform based on gesture
  recognition technology.
\newblock In {\em 2021 China Automation Congress (CAC)}, pages 940--944, 2021.

\bibitem{Galluzzi2015}
Valerie Galluzzi, Ted Herman, and Philip Polgreen.
\newblock {Hand hygiene duration and technique recognition using wrist-worn
  sensors}.
\newblock In {\em Proceedings of the 14th International Conference on
  Information Processing in Sensor Networks - IPSN '15}, pages 106--117, New
  York, New York, USA, 2015. ACM Press.

\bibitem{bal2017}
Mert Bal and Reza Abrishambaf.
\newblock A system for monitoring hand hygiene compliance based-on
  internet-of-things.
\newblock In {\em 2017 IEEE International Conference on Industrial Technology
  (ICIT)}, pages 1348--1353. IEEE, 2017.

\bibitem{Li2018}
Hong Li, Shishir Chawla, Richard Li, Sumeet Jain, Gregory~D. Abowd, Thad
  Starner, Cheng Zhang, and Thomas Plotz.
\newblock {WristWash: Towards automatic handwashing assessment using a
  wrist-worn device}.
\newblock {\em Proceedings - International Symposium on Wearable Computers,
  ISWC}, pages 132--139, 2018.

\bibitem{jain2018}
Ankita Jain and Vivek Kanhangad.
\newblock Gender classification in smartphones using gait information.
\newblock {\em Expert Systems with Applications}, 93:257--266, 2018.

\bibitem{van2019}
Tim Van~Hamme, Giuseppe Garofalo, Enrique Argones~R{\'u}a, Davy Preuveneers,
  and Wouter Joosen.
\newblock A systematic comparison of age and gender prediction on imu
  sensor-based gait traces.
\newblock {\em Sensors}, 19(13):2945, 2019.

\bibitem{Mondol2015}
Md~Abu~Sayeed Mondol and John~A. Stankovic.
\newblock {Harmony: A Hand Wash Monitoring and Reminder System using Smart
  Watches}.
\newblock In {\em Proceedings of the 12th EAI International Conference on
  Mobile and Ubiquitous Systems: Computing, Networking and Services}. ACM,
  2015.

\bibitem{mondol2020}
Md~Abu~Sayeed Mondol and John~A Stankovic.
\newblock Hawad: Hand washing detection using wrist wearable inertial sensors.
\newblock In {\em 2020 16th International Conference on Distributed Computing
  in Sensor Systems (DCOSS)}, pages 11--18. IEEE, 2020.

\bibitem{banerjee2020}
Ayan Banerjee, Venkata Naga Sai~Apurupa Amperyani, and Sandeep~KS Gupta.
\newblock Hand hygiene compliance checking system with explainable feedback.
\newblock In {\em Proceedings of the 6th ACM Workshop on Wearable Systems and
  Applications}, pages 34--36, 2020.

\bibitem{Samyoun2021}
Sirat Samyoun, Sudipta~Saha Shubha, Md~Abu {Sayeed Mondol}, and John~A.
  Stankovic.
\newblock {iWash: A smartwatch handwashing quality assessment and reminder
  system with real-time feedback in the context of infectious disease}.
\newblock {\em Smart Health}, 19(January):100171, mar 2021.

\bibitem{wu2021}
Fan Wu, Taiyang Wu, David~Cheng Zarate, Richard Morfuni, Bronte Kerley, Jason
  Hinds, David Taniar, Mark Armstrong, and Mehmet~Rasit Yuce.
\newblock An autonomous hand hygiene tracking sensor system for prevention of
  hospital associated infections.
\newblock {\em IEEE Sensors Journal}, 21(13):14308--14319, 2021.

\bibitem{wahl2022}
Karina Wahl, Philipp~Marcel Scholl, Silvan Wirth, Marcel Miché, Jeannine
  Häni, Pia Schülin, and Roselind Lieb.
\newblock On the automatic detection of enacted compulsive hand washing using
  commercially available wearable devices.
\newblock {\em Computers in Biology and Medicine}, 143:105280, 2022.

\bibitem{fagert2022}
Jonathon Fagert, Amelie Bonde, Sruti Srinidhi, Sarah Hamilton, Pei Zhang, and
  Hae~Young Noh.
\newblock Clean vibes: Hand washing monitoring using structural vibration
  sensing.
\newblock {\em ACM Trans. Comput. Healthcare}, jan 2022.

\bibitem{fix1989}
Evelyn Fix and Joseph~Lawson Hodges.
\newblock Discriminatory analysis. nonparametric discrimination: Consistency
  properties.
\newblock {\em International Statistical Review/Revue Internationale de
  Statistique}, 57(3):238--247, 1989.

\bibitem{altman1992}
Naomi~S Altman.
\newblock An introduction to kernel and nearest-neighbor nonparametric
  regression.
\newblock {\em The American Statistician}, 46(3):175--185, 1992.

\bibitem{naftaly1997}
Ury Naftaly, Nathan Intrator, and David Horn.
\newblock Optimal ensemble averaging of neural networks.
\newblock {\em Network: Computation in Neural Systems}, 8(3):283, 1997.

\bibitem{breiman1996}
Leo Breiman.
\newblock Bagging predictors.
\newblock {\em Machine learning}, 24(2):123--140, 1996.

\bibitem{freund1996}
Yoav Freund, Robert~E Schapire, et~al.
\newblock Experiments with a new boosting algorithm.
\newblock In {\em icml}, volume~96, pages 148--156. Citeseer, 1996.

\bibitem{ho1998}
Tin~Kam Ho.
\newblock Nearest neighbors in random subspaces.
\newblock In {\em Joint IAPR International Workshops on Statistical Techniques
  in Pattern Recognition (SPR) and Structural and Syntactic Pattern Recognition
  (SSPR)}, pages 640--648. Springer, 1998.

\bibitem{li2011}
Shengqiao Li, E~James Harner, and Donald~A Adjeroh.
\newblock Random knn feature selection-a fast and stable alternative to random
  forests.
\newblock {\em BMC bioinformatics}, 12(1):1--11, 2011.

\bibitem{steinwart2008}
Ingo Steinwart and Andreas Christmann.
\newblock {\em Support vector machines}.
\newblock Springer Science \& Business Media, 2008.

\bibitem{Albawi2017}
S.~{Albawi}, T.~A. {Mohammed}, and S.~{Al-Zawi}.
\newblock Understanding of a convolutional neural network.
\newblock In {\em 2017 International Conference on Engineering and Technology
  (ICET)}, pages 1--6, 2017.

\bibitem{hochreiter1997}
Sepp Hochreiter and J{\"u}rgen Schmidhuber.
\newblock Long short-term memory.
\newblock {\em Neural computation}, 9(8):1735--1780, 1997.

\bibitem{yu2019}
Yong Yu, Xiaosheng Si, Changhua Hu, and Jianxun Zhang.
\newblock A review of recurrent neural networks: Lstm cells and network
  architectures.
\newblock {\em Neural computation}, 31(7):1235--1270, 2019.

\bibitem{leotta2021}
Maurizio Leotta, Andrea Fasciglione, and Alessandro Verri.
\newblock Daily living activity recognition using wearable devices: A
  features-rich dataset and a novel approach.
\newblock In Alberto Del~Bimbo, Rita Cucchiara, Stan Sclaroff, Giovanni~Maria
  Farinella, Tao Mei, Marco Bertini, Hugo~Jair Escalante, and Roberto Vezzani,
  editors, {\em Pattern Recognition. ICPR International Workshops and
  Challenges}, pages 171--187, Cham, 2021. Springer International Publishing.

\bibitem{leotta2021data}
Maurizio Leotta, Andrea Fasciglione, and Alessandro Verri.
\newblock {Daily Living Activity Recognition Using Wearable Devices: A
  Features-rich Dataset and a Novel Approach}, 2021.

\bibitem{Shimmer3}
Shimmer.
\newblock Shimmer3 imu unit, 2019.
\newblock Online accessed 26 July 2021.

\bibitem{Hjorth1970}
Bo~Hjorth.
\newblock Eeg analysis based on time domain properties.
\newblock {\em Electroencephalography and Clinical Neurophysiology}, 29(3):306
  -- 310, 1970.

\bibitem{kim2004}
Tae-Hwan Kim and Halbert White.
\newblock On more robust estimation of skewness and kurtosis.
\newblock {\em Finance Research Letters}, 1(1):56--73, 2004.

\bibitem{Wang2015}
Z.~Wang and T.~Oates.
\newblock Imaging time-series to improve classification and imputation.
\newblock In {\em IJCAI International Joint Conference on Artificial
  Intelligence}, volume 2015-January, pages 3939--3945, 2015.

\bibitem{baldini2017}
Gianmarco Baldini, Gary Steri, Raimondo Giuliani, and Claudio Gentile.
\newblock Imaging time series for internet of things radio frequency
  fingerprinting.
\newblock In {\em 2017 International Carnahan Conference on Security Technology
  (ICCST)}, pages 1--6. IEEE, 2017.

\bibitem{qin2020}
Zhen Qin, Yibo Zhang, Shuyu Meng, Zhiguang Qin, and Kim-Kwang~Raymond Choo.
\newblock Imaging and fusing time series for wearable sensor-based human
  activity recognition.
\newblock {\em Information Fusion}, 53:80--87, 2020.

\bibitem{ioffe2015}
Sergey Ioffe and Christian Szegedy.
\newblock Batch normalization: Accelerating deep network training by reducing
  internal covariate shift.
\newblock In {\em International conference on machine learning}, pages
  448--456. PMLR, 2015.

\bibitem{sokolova2009}
Marina Sokolova and Guy Lapalme.
\newblock A systematic analysis of performance measures for classification
  tasks.
\newblock {\em Information processing \& management}, 45(4):427--437, 2009.

\bibitem{Fagerland2013}
Morten~W Fagerland, Stian Lydersen, and Petter Laake.
\newblock The mcnemar test for binary matched-pairs data: mid-p and asymptotic
  are better than exact conditional.
\newblock {\em BMC medical research methodology}, 13(1):91, 2013.

\bibitem{liu2005}
Huan Liu and Lei Yu.
\newblock Toward integrating feature selection algorithms for classification
  and clustering.
\newblock {\em IEEE Transactions on knowledge and data engineering},
  17(4):491--502, 2005.

\end{thebibliography}

\begin{IEEEbiography}[{\includegraphics[width=1in,height=1.25in,clip,keepaspectratio]{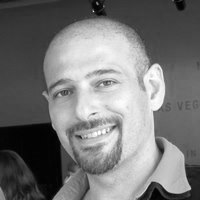}}]{EMANUELE LATTANZI} 
received the Laurea degree (summa cum laude) in 2001 and the Ph.D. degree from the University of Urbino, Italy, in 2005. Since 2001, he has been with the Information Science and Technology Institute, University of Urbino. In 2003, he was with the Department of Computer Science and Engineering, Pennsylvania State University, as a Visiting Scholar with Prof. V. Narayanan. From  2008 until 2020, he has been Assistant Professor in Computer Engineering at the Department of Pure and Applied Sciences (DiSPeA) of the University of Urbino, Italy, where hi is currently Associate Professor in Computer Engineering. His research interests include wireless sensor networks, energy-aware routing algorithms, artificial intelligence, and Internet of Things.
\end{IEEEbiography}

\begin{IEEEbiography}[{\includegraphics[width=1in,height=1.25in,clip,keepaspectratio]{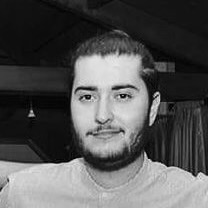}}]{LORENZO CALISTI} received the Laurea degree in Applied Computer Science from the University of Urbino, Italy, in 2020. He is currently a graduating student at the University of Urbino. His research interests include machine learning, embedded systems.
\end{IEEEbiography}

\begin{IEEEbiography}[{\includegraphics[width=1in,height=1.25in,clip,keepaspectratio]{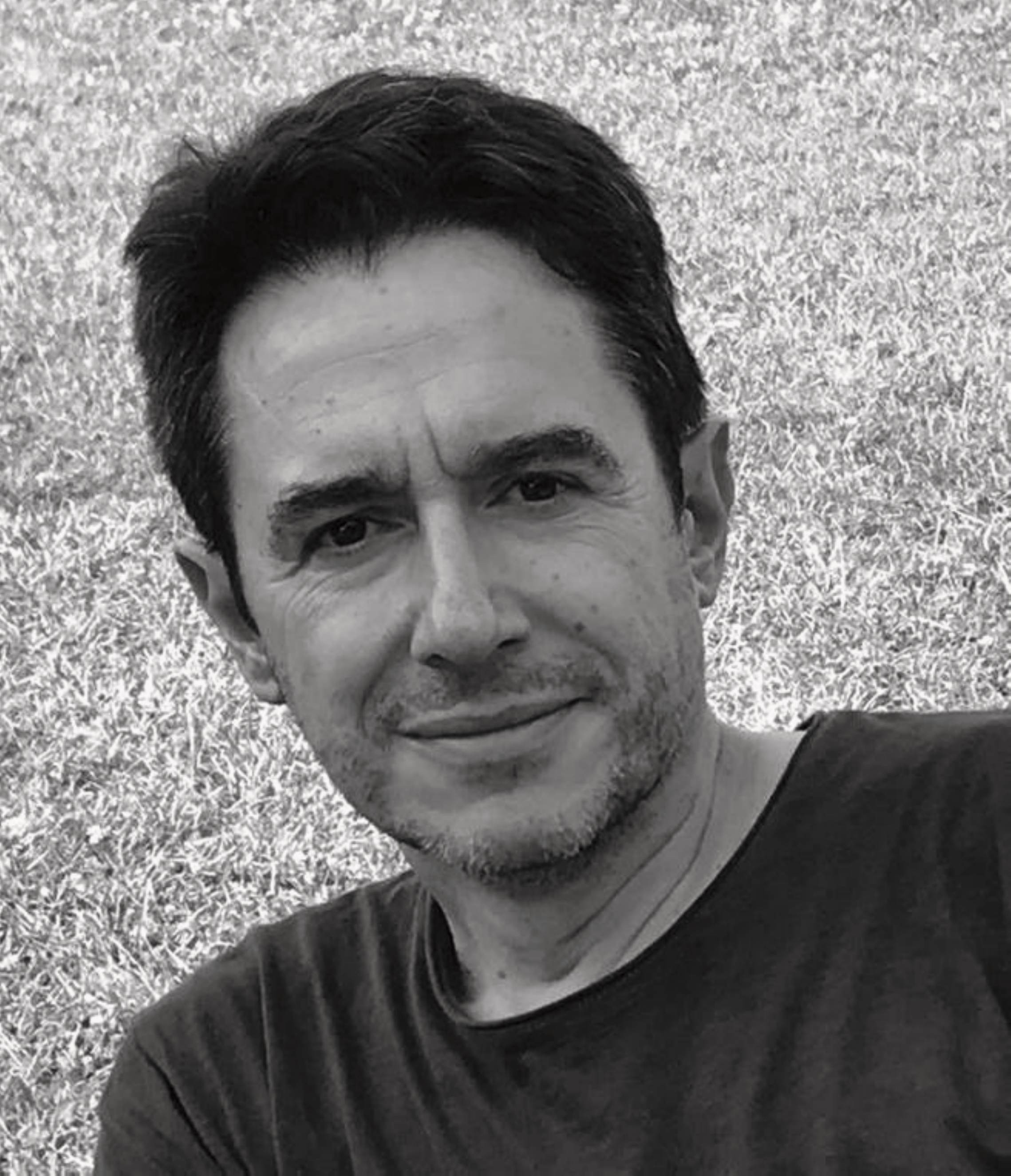}}]
{VALERIO FRESCHI}
graduated in Electronic Engineering at University of Ancona, Italy, in 1999 and received the Ph.D. degree in Computer Science Engineering from University of Ferrara, Italy in 2006. He is currently Assistant Professor in Computer Engineering at the Department of Pure and Applied Sciences (DiSPeA) of the University of Urbino, Italy. His research interests include wireless sensor networks, energy-efficient algorithms, optimization, machine learning.
\end{IEEEbiography}

\EOD

\end{document}